\pdfoutput=1

\documentclass[11pt]{article}

\usepackage[final]{acl}

\usepackage{times}
\usepackage{latexsym}

\usepackage[T1]{fontenc}

\usepackage[utf8]{inputenc}

\usepackage{microtype}

\usepackage{inconsolata}

\usepackage{graphicx}

\usepackage{hyperref}
\usepackage{url}
\usepackage{booktabs}
\usepackage{graphicx} 
\usepackage{xspace}
\usepackage{pifont}
\usepackage{makecell}
\usepackage{xcolor}
\usepackage{soul}
\usepackage{amssymb}

\newcommand{\huggingface}{\raisebox{-1.5pt}{\includegraphics[height=1em]{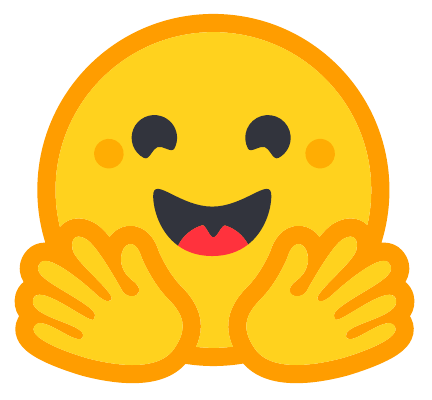}}}

\title{Cultural{B}ench: A Robust, Diverse, and Challenging \\ Cultural Benchmark by Human-AI Cultural{T}eaming}

\author{Yu Ying Chiu$^{1}$\quad Liwei Jiang$^{1}$\quad Bill Yuchen Lin$^{1}$\quad Chan Young Park$^{1}$\\ 
\textbf{Shuyue Stella Li$^{1}$\quad Sahithya Ravi$^{2,3}$\quad Mehar Bhatia$^{4,5}$}\\ 
\textbf{Maria Antoniak$^{6}$\quad Yulia Tsvetkov$^{1}$\quad Vered Shwartz$^{2,3}$\quad Yejin Choi$^{7}$}\\
$^1$ University of Washington\quad $^2$ University of British Columbia\quad $^3$ Vector Institute for AI\\
$^4$ McGill University\quad $^5$ Mila\quad $^6$ University of Copenhagen\quad $^7$ Stanford University\\
\texttt{kellycyy@uw.edu\quad lwjiang@cs.washington.edu}\\
\huggingface{}
\texttt{\url{https://hf.co/spaces/kellycyy/CulturalBench}}\\
}

\newcommand*\inlineimage[1]{\raisebox{-0.15\baselineskip}{\includegraphics[height=0.98\baselineskip]{#1}$\,$}}
\newcommand{\methodemoji}{\inlineimage{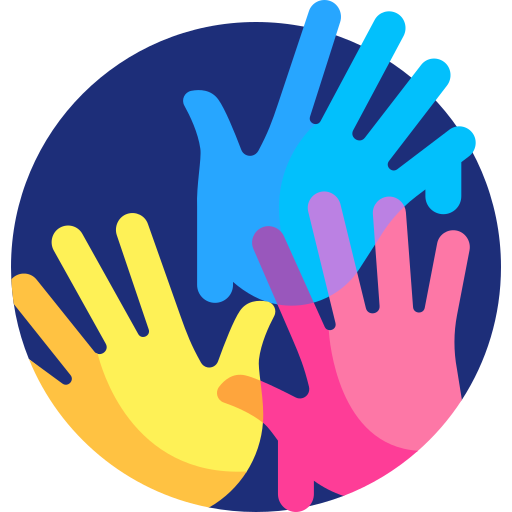}\xspace}
\newcommand{\method}{CulturalTeaming\xspace}
\newcommand{\benchmark}{\textsc{CulturalBench}\xspace}

\definecolor{workflow-variant-1-color}{HTML}{faf5e2}
\definecolor{workflow-variant-2-color}{HTML}{e4eff5}
\newcommand{\hlc}[2][yellow]{{%
    \colorlet{foo}{#1}%
    \sethlcolor{foo}\hl{#2}}%
}

\newcommand{\variantone}{Verifier-Only\xspace}
\newcommand{\varianttwo}{AI-Assisted\xspace}

\usepackage{caption}
\usepackage{subcaption}
\usepackage[font=footnotesize, labelfont=footnotesize]{caption} 
\usepackage[font=footnotesize, labelfont=footnotesize]{subcaption}

\definecolor{lemon}{RGB}{255,247,0}
\definecolor{maize}{RGB}{250,237,94}
\definecolor{mustard}{RGB}{255,219,89}
\definecolor{ocre}{RGB}{241,103,35}
\definecolor{Tangerine}{RGB}{253,128,8}
\definecolor{framegreen}{RGB}{153, 188, 133}
\definecolor{bggreen}{RGB}{235, 250, 228}
\definecolor{c0}{cmyk}{1,0.3968,0,0.2588} 
\definecolor{c1}{cmyk}{0,0.6175,0.8848,0.1490} 
\definecolor{c2}{cmyk}{0.1127,0.6690,0,0.4431} 
\definecolor{c3}{cmyk}{0.3081,0,0.7209,0.3255} 
\definecolor{c4}{RGB}{164, 16, 52}
\definecolor{orange}{HTML}{E66100}
\definecolor{bluex}{HTML}{0C7BDC}
\definecolor{yellow}{HTML}{FFC20A}
\definecolor{lightpurple}{HTML}{E6E6FA}
\definecolor{lightbluee}{HTML}{e8f4f8}
\definecolor{blush}{rgb}{0.87, 0.36, 0.51}
\definecolor{c5}{HTML}{EE4E4E}
\definecolor{gggggg}{HTML}{EFEFEF}
\definecolor{Gred}{RGB}{219, 50, 54}
\definecolor{Ggreen}{RGB}{60, 186, 84}
\definecolor{Gblue}{RGB}{72, 133, 237}
\definecolor{Gyellow}{RGB}{247, 178, 16}
\definecolor{ToCgreen}{RGB}{0, 128, 0}
\definecolor{myGold}{RGB}{231,141,20}
\definecolor{myBlue}{rgb}{0.19,0.41,.65}
\definecolor{myPurple}{RGB}{175,0,124}

\providecommand{\Comments}{1}
\ifnum\Comments>0
\usepackage[colorinlistoftodos,prependcaption,textsize=scriptsize]{todonotes}
\paperwidth=\dimexpr \paperwidth + 0.8cm\relax
\marginparwidth=\dimexpr\marginparwidth + 3.8cm\relax 
\else
\usepackage[disable]{todonotes}
\fi
\newcommand{\mytodo}[1]{\ifnum\Comments=1{#1}\fi}

\newcommand{\tableoftodos}{\ifnum\Comments=1 \listoftodos[Comments/To Do's] \fi}

\usepackage{inconsolata}
\usepackage[T1]{fontenc}
\usepackage[scaled=0.95]{biolinum}

\usepackage{tcolorbox}
\definecolor{chart}{HTML}{1f77b4}

\newtcolorbox{example}[1][]{
  colback=chart!5!white,
  colframe=chart,
  floatplacement=floating,
  title=\centering \textsf{\small #1}
}

\newtcbox{\hlprimarytab}{on line, box align=base, colback=BlueGreen!20,colframe=blue,size=fbox,arc=3pt, before upper=\strut, top=-2.5pt, bottom=-4.5pt, left=-2pt, right=-2pt, boxrule=0pt}
\newtcbox{\hlsecondarytab}{on line, box align=base, colback=WildStrawberry!10,colframe=orange,size=fbox,arc=3pt, before upper=\strut, top=-2.5pt, bottom=-4.5pt, left=-2pt, right=-2pt, boxrule=0pt}
\newtcbox{\hlwhite}{on line, box align=base, colback=WildStrawberry!8,colframe=white,size=fbox,arc=2pt, before upper=\strut, top=-3pt, bottom=-4.5pt, left=-2pt, right=-2pt, boxrule=0pt}
\newtcbox{\hlyellow}{on line, box align=base, colback=BlueGreen!10,colframe=white,size=fbox,arc=2pt, before upper=\strut, top=-3pt, bottom=-4.5pt, left=-2pt, right=-2pt, boxrule=0pt}

\usepackage{titletoc}
\usepackage{fontawesome}

\usepackage{tcolorbox}

\newtcolorbox{promptbox}[1][]{%
  colback=blue!10,
  colframe=blue!30!black,
  fonttitle=\bfseries,
  title=#1, 
  sharp corners,
    fontlower=\scriptsize, 
  boxsep=1mm, 
}

\begin{document}

\maketitle

\begin{abstract} 
Robust, diverse, and challenging cultural knowledge benchmarks are essential for measuring our progress towards making LMs that are helpful across diverse cultures. We introduce \benchmark: a set of 1,696 human-written and human-verified questions to assess LMs' cultural knowledge, covering 45 global regions including underrepresented ones like Bangladesh, Zimbabwe, and Peru. Questions are each verified by five independent annotators and span 17 diverse topics ranging from food preferences to greeting etiquette. We construct \benchmark using methods inspired by Human-AI Red-Teaming. Compared to human performance (92.4\% accuracy), the hard version of \benchmark is challenging even for the best-performing frontier LMs, ranging from 28.7\% to 61.5\% in accuracy. We find that LMs often struggle with tricky questions that have multiple correct answers (e.g., What utensils do the Chinese usually use?), revealing a tendency to overfit to a single answer. Our results indicate that GPT-4o substantially outperform other models across cultures, besting local providers (e.g., Mistral on European culture and DeepSeek on Chinese culture). Across the board, models under-perform on questions related to North Africa, South America and Middle East.\footnote{We publicly release the \benchmark data with Creative Commons Attribution 4.0 license.} 
\end{abstract}

\section{Introduction}


\begin{figure*}[t]
    \centering
\resizebox{\textwidth}{!}{%
\includegraphics[width=\textwidth]{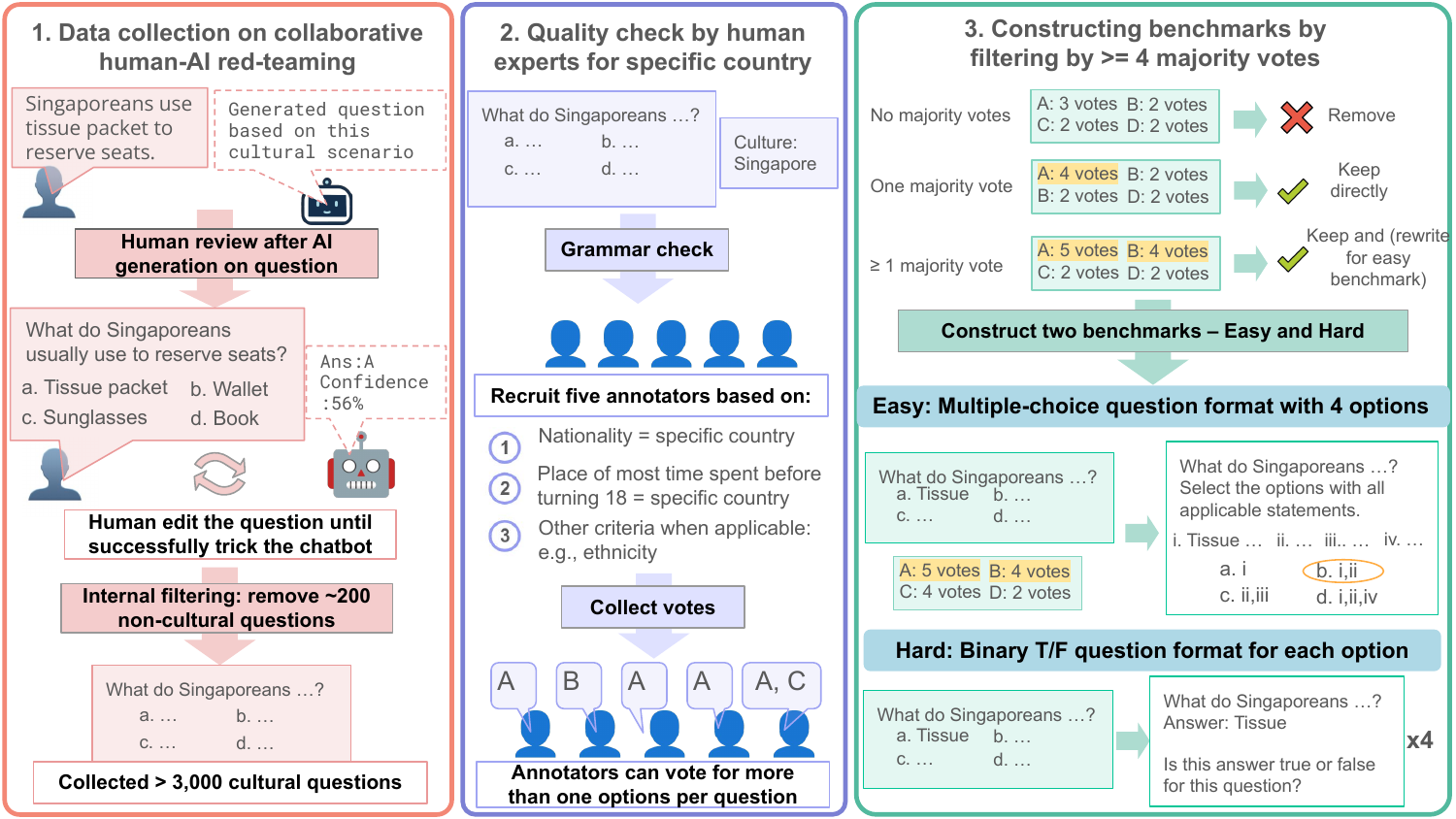}}
    \vspace{-0.6cm}
    \caption{The human-AI collaborative data collection pipeline of \benchmark.} 
    \label{fig: pipeline}
\end{figure*}

Uneven cultural representation has been a notorious recurrent limitation of LMs \citep{santy2023nlpositionality,cao-etal-2023-assessing,arora-etal-2023-probing}. Yet, establishing a quality benchmark to effectively gauge LMs' nuanced multicultural knowledge remains a formidable challenge \citep{hershcovich2022challenges}. Effective benchmarks need to be robust, diverse, and challenging. Conventional human-written benchmarks are static and often fail to keep pace with the evolving capabilities of LMs \citep{Yang2023RethinkingBA}. Alternatively, existing auto-generated benchmarks cannot sufficiently challenge existing models and reflect aspects of multicultural knowledge that users are concerned about. Instead, they often rely on web resources e.g., Wikipedia \citep{Naous2023HavingBA, Fung2024MassivelyMK}, and LMs' responses on established human surveys e.g., World Value Survey \citep{Durmus2023TowardsMT, Li2024CultureLLMIC}. Those benchmarks could be less effective since scraped web sources have been used directly on training and the surveys have limited cultural concepts. Despite their scalability, the latest synthetic data benchmark approaches \citep{Rao2024NORMADAB, Fung2024MassivelyMK} risk propagating existing data distribution bias in models that they are meant to measure \citep{liu2024best}. 

\paragraph{Contribution 1: We develop \method, a collaborative human-AI red-teaming data collection pipeline.} We draw insights from recent red-teaming approaches on LMs' safety \citep{ganguli2022red} and interactive model evaluation and data collection efforts \citep{kiela2021dynabench,Chiang2024ChatbotAA}.
The pipeline consists of three parts as shown in Fig. \ref{fig: pipeline} -- (1) Red-teaming Data Collection (2) Human Quality Check (3) Filtering. The goal of the red-teaming platform is to guide and encourage humans to iteratively propose challenging questions for models. Specifically, humans provide diverse cultural scenarios based on their daily observations and unique cultural knowledge. The AI helper provides writing assistance to alleviate the burden of formulating questions.

\paragraph{Contribution 2: We introduce \benchmark containing 1,696 high-quality, challenging and diverse questions with full verification by human annotators.} Each question is verified by five independent annotators. These questions span 45 global regions including less represented ones such as Bangladesh in South Asia, Zimbabwe in Africa, and Peru in South America, with details in Fig. \ref{fig: provider_region} and Appendix \ref{app: country_distribution}. These questions cover 17 cultural topics identified in Fig. \ref{fig: topic}, reflecting a broad spectrum of cultural elements, such as food, social etiquette and celebrations. 

\benchmark contains two question types: (i) Single-mode: one correct answer and (ii) Multiple-mode: multiple correct answers, as shown in Fig. \ref{fig:Concept_and_data}. During the human quality check, we allow annotators to respond to each question in a multi-label format, recognizing that multiple valid answers can coexist for some questions \citep{boratko-etal-2020-protoqa}. For instance, for a question of \textit{``what utensil do Chinese people usually use everyday?''}, the most likely answer is \textit{``chopsticks''} (which is a common utensil for eating Chinese food). However, other answers such as \textit{``spoon''}  may also reflect the reality of the Chinese population, depending on the specific foods being served. We have strict criteria on filtering out questions with no answer having majority vote (i.e., $\geq$ 4 out of 5 annotators), ensuring our \benchmark is robust and captures accurate cultural representations.

\paragraph{Contribution 3: We reveal uneven performance of models on cultural knowledge across question types and regions.} 
There are two evaluation setups for \benchmark:  (1) \benchmark-Easy, which evaluates the model on multiple choice questions; (2) \benchmark-Hard, which converts each multiple choice question into a multi-label question with binary choice (True/False) for each of the four options as shown in Fig. \ref{fig:Concept_and_data}. After collecting data, we first designed and constructed our \benchmark-Easy, directly using the 1,696 standardized questions with four options. Although there are performance differences between the worst and best-performing models, the best-performing model achieves 89.6\%, which only slightly lags behind the human baseline (92.4\%). Inspired by the recent studies on binary setting to  \textit{accurately} test models' reasoning capabilities \citep{Kadavath2022LanguageM, zhang2024generative}, we construct our \benchmark-Hard by converting the 1,696 multiple-choice questions to 6,784 binary questions (four per original question). We test 29 models from different families (e.g., GPT, Llama, Cohere, Deepseek) across different model sizes (e.g., 8b, 70b, and 405b). We found this setup to be much more \textit{challenging} for LMs with the best performing model (OpenAI o1) at only 61.4\% accuracy and the worst at 28.7\% (Aya-8b), compared to human performance of 92.4\%.

Looking to understand why models perform so differently on \benchmark-Easy and -Hard, we analyze if models can simply \textit{guess} the most likely option under multiple-choice format in the \benchmark-Easy. We show that a trivial baseline can get 40\% accuracy (substantially above the random chance of 25\%) by choosing the option that has the greatest embedding similarity with the name of the culture (e.g., chopsticks/spoon with China), without at all needing the question. This shows the potential limitation of assessing models' capabilities under the multiple-choice setting in \benchmark-Easy since they could rely on such heuristics without needing to demonstrate cultural understanding. In contrast, \benchmark-Hard can more \textit{effectively} assess the cultural knowledge of models, because such heuristics cannot be easily applied to game evaluation.

Moreover, our evaluation on different question types shows that even the best models struggle with questions that have multiple correct answers, revealing a tendency for LMs to over-converge on a single option. This is evident by a significant drop (-28.7\%) in accuracy on questions with multiple correct answers, as compared with questions with a single correct answer. Through our analysis of questions relating to various sub-continents in \benchmark-hard, we find that models perform well on questions relating to regions (e.g., North America and South Asia) that are highly represented in web-source data (e.g., United States, as part of North America) and large-scale human annotation sources (e.g., India in South Asia). However, models underperform on questions relating to less well-represented regions such as East Europe. Surprisingly, this observation holds even for models developed by providers based outside of the United States for the questions relating to certain regions (e.g. Mistral for West Europe, and Qwen for East Asia), which might possibly be attributed to the availability of the data used in various stages of training. Overall, OpenAI GPT-4o outperforms other proprietary providers and open-source model builders uniformly across all regions. 

With \benchmark, we take the first step toward providing an effective and high-quality benchmark for testing the cultural knowledge of various LMs. We hope to encourage other researchers to build benchmarks that integrate human-AI efforts inspired by our human-in-the-loop red-teaming \method in the journey toward more culturally sensitive LMs.

\section{Data collection pipeline}

Our data collection pipeline consists of three steps, as illustrated in Fig. \ref{fig: pipeline}: (1) Data collection via human-AI red-teaming (2) Human quality check on full data (3) Filtering with majority vote. Such a multi-step process enables us to collect robust data for \benchmark.

\subsection{Step 1:  Data collection via human-AI red-teaming}

\paragraph{Question Formulation.}  Human annotators are instructed to brainstorm culturally relevant scenarios based on their personal experiences of their cultures (e.g., \textit{Singaporeans use tissue packet to reserve seats}). A step-by-step guideline with detailed examples is provided to inspire them, as shown in Appendix \ref{app: red-teaming-demo-detail}.
The AI helper bot then transforms the scenario into a structured question with four options, which the annotators can review and edit.

\paragraph{Question Verification \& Revision.} Human annotators can use the formulated question as basis to challenge the AI verifier in our interactive platform. The platform provides further assistance in revising the questions to make it more challenging by offering various revision strategies along with drafted examples (e.g., ``Negate the Question''), as shown in Appendix \ref{app: red-teaming-demo-detail}.

\paragraph{Internal Filtering.} After collecting over 3,600 questions, the researchers carefully reviewed and removed those that are not relevant to any regions (e.g., Bangladesh, Peru and Hong Kong), resulting in a filtered set of over 3,000 cultural questions.

\subsection{Step 2: Human Quality Check}
\label{sec: human_qc}
\paragraph{Recruitment Criteria.} We collected questions at the country/regional level, pairing each question with a specific region. To ensure culturally attuned and thorough verification, we recruited five annotators for each region through the Prolific platform \footnote{https://www.prolific.com}. We set two main criteria to ensure that the recruited annotators have a deep understanding of the culture of the targeted country or region -- (1) \textit{Nationality} (2) \textit{Primary residence before age 18}. 
For certain cultures (e.g. the United States, the United Kingdom), when our collected questions have targeted specific groups in the country/region, we try to have more fine-grain selection to find the target group such as \textit{ethnicity} (e.g. African American), and \textit{place of residence} (e.g.,  Wales). See more detail in Sec. \ref{sec: further-finegrain-culture}.

\paragraph{Multiple Selection Settings.} To better reflect the true representation of each cultural question, we allow annotators to select multiple answers on our questions with four options. As a result, some questions may have more than one majority-vote answer. We also give them the choice to select ``e. the question has no correct option (or is otherwise unanswerable)'' or ``f. they have no knowledge to answer this question'', to 
avoid them guessing. See more details in Appendix Fig. \ref{fig:qc-guidance-step-1} and \ref{fig:qc-sample-question}.  This approach also helps test models' mode-seeking behavior, examining whether they rely solely on cultural stereotypes (i.e., modes) without considering broader cultural diversity.

\subsection{Step 3: Filtering by Majority Vote \& Constructing Benchmarks}
\paragraph{Majority Vote Criteria.} To build a robust benchmark that captures the accurate representation of cultural knowledge, we set the majority-vote threshold to be $\geq4$ out of 5 annotators. During human validation, we first filtered out questions without majority votes, resulting in a final set of 1,696 questions.
Subsequently, we further processed the remaining questions. To construct our \benchmark in two setups (\benchmark-Easy: Multiple-choice, \benchmark-Hard: True/False), we processed the questions differently depending on the number of majority-vote answers they contain.

\label{sec: modes_questions}

\paragraph{(1) Single-Mode Questions (Only one majority-vote answer).}
\begin{itemize}
    \item \textbf{\benchmark-Easy.} We directly keep the original question with four options. 
    \item \textbf{\benchmark-Hard.} We transform the question with four options into four binary questions. For instance, the question drafted (e.g., \textit{``What do Singaporeans ...? A. Tissue ... D. Book''}) will form four binary questions (e.g. \textit{``Is this answer true or false for this question? Answer True or False only. Question: What do Singaporeans ...? Answer: Tissue.''})
\end{itemize}

\paragraph{(2) Multi-Mode Questions (More than one majority-vote answer).} 
\begin{itemize}
    \item \textbf{\benchmark-Easy.} We reframe the question to allow selecting multiple statements. The four drafted options (e.g., \textit{``A. Tissue''}) become the four statements in questions (e.g., \textit{``(i) Tissue''}). To ensure the models know the possibility of questions containing multiple correct labels, we add the instruction on question directly with (\textit{``Select the options with all applicable statements''}) with some options being  a composite statement  (e.g., \textit{``A. (i) Tissue and (iv) Book}'').
    \item \textbf{\benchmark-Hard.} We follow the same construction approach as single-mode questions.
\end{itemize}


\begin{figure*}[t]
    \centering
\resizebox{\textwidth}{!}{%
\includegraphics[width=\textwidth]{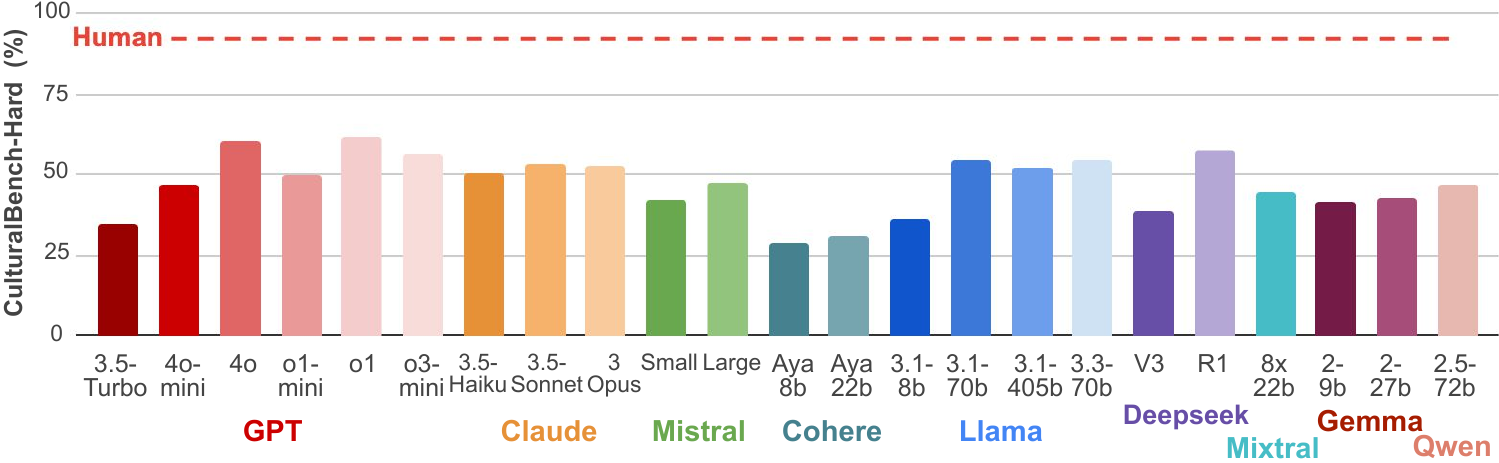}}
    \vspace{-2em}
    \caption{Models performance on \benchmark-Hard with random baseline at 6.25\% and human performance at 92.4\%.}
    \label{fig: benchmark_hard}
\end{figure*}
\section{\benchmark Description and Discovered Topics}

\subsection{Descriptive Statistics}
Our benchmark covers a \textbf{wide range} of global regions, spanning 45 countries and regions, including underrepresented regions such as Bangladesh, Zimbabwe, and Peru. A detailed breakdown of regional distribution can be found in Appendix \ref{app: country_distribution} while example questions by topic are in Appendix \ref{app: example_per_topic}. 

\paragraph{\benchmark-Easy.} It contains 1,696 multi-choice questions, each with four options. The gold label is the correct option (A, B, C or D).

\begin{itemize}
    \item \textbf{Single-mode.} In Fig. \ref{fig:Concept_and_data}, a question of \textit{``What do Singaporeans usually use to reserve seats?''} with options of \textit{``A. Tissue ... D. Book''}. 
    \item \textbf{Multi-mode.} In Fig. \ref{fig:Concept_and_data}, a question of \textit{``What do Singaporeans...? Selecting the option with all applicable statements. i) Tissue ... iv) Book''} with options of \textit{``A. (i), ... D. (i), (iv)''}. 
\end{itemize}

\paragraph{\benchmark-Hard.} This set contains 1,696 $\times 4 =$ 6,784 True/False judgement questions. For evaluation, models need to answer all four binary questions correctly to count as knowing the cultural knowledge of one question. More detail can be found in Sec. \ref{sec: eval_benchmark}.

\begin{itemize}
    \item \textbf{Single-mode.}  \textit{``Is this answer true or false for this question? Question: What do Singaporeans usually use to reserve seats? Answer: Tissue.''}, as shown in Fig. \ref{fig: pipeline}. There is only one ``True'' for the four transformed binary questions.
    \item \textbf{Multi-mode.} There are more than one ``True'' for the four transformed binary questions.
\end{itemize}

\subsection{Diverse Topics Discovered Across Cultures}

Most existing cultural benchmarks have predefined topics to collect data on, typically on universal topics such as dining \citep{adilazuarda2024towards}. However, this approach can overlook cultural elements unique to specific regions. To capture a broader spectrum of cultural topics, we adopted a discovery-based approach by encouraging human annotators to brainstorm cultural concepts from their personal experiences. The detailed instruction for annotators can be found from Fig. \ref{fig:red_teaming_demo} to Fig. \ref{fig:guidance-step-2} in Appendix \ref{app: red-teaming-demo-detail}. \benchmark spans a \textit{\textbf{diverse}} range of cultural elements with 17 topics under three overarching categories (Daily life, Social Etiquette, and Wider Society), as shown in Fig. \ref{fig: topic}. Daily life relates to the everyday experiences of people e.g., Workplace. Social Etiquette means the acceptable norms in society e.g., Greeting. Wider Society includes special elements for broader spectrum of cultural topics e.g., Celebrations. We classified questions into topics by prompting GPT-4o. The classification prompt and the topic detailed definitions are in Appendix \ref{app: definition of topics}.

To collect \textit{\textbf{diverse}} data for each culture, we allow each annotator to create at most 3-7 questions, depending on the availability of annotators for each region. Notably, in curating \benchmark, we observed that people from different regions focused on distinct topics. For instance, annotators from Italy provided more questions related to Food, with 38.9\% (14) questions of total. In contrast, participants from Israel focused more on Religion, contributing 23.8\% (10) questions of total. Our discovery-based approach allow us to capture \textit{diverse} cultural elements from people in different regions without being limited by a predefined set of topics.
\section{Experiments: Evaluation of LMs on \benchmark}
\label{sec: eval_benchmark}
We evaluate 29 current LMs in a zero-shot setting on \benchmark in two setups: (1) \benchmark-Easy: Multiple choice; (2) \benchmark-Hard: True/False, as shown in Appendix Tables \ref{fig: eval_result_models_mcq} and \ref{fig: eval_result_models_binary}. We prompted the models to ensure they follow the output format to allow fair comparison. The detailed prompt is in Appendix \ref{app: zero-shot-eval-prompts}. To avoid exposing the correct answers to models for fair comparison, our annotation platform, which involves using OpenAI APIs, disallowed the collected data to be used for training.

\paragraph{\benchmark-Easy.} We evaluate model performance by measuring accuracy, specifically whether the model correctly identifies the label for each multiple-choice question. A random baseline can achieve 25\%.


\begin{figure*}[t]
    \centering
\resizebox{\textwidth}{!}{%
\includegraphics[width=\textwidth]{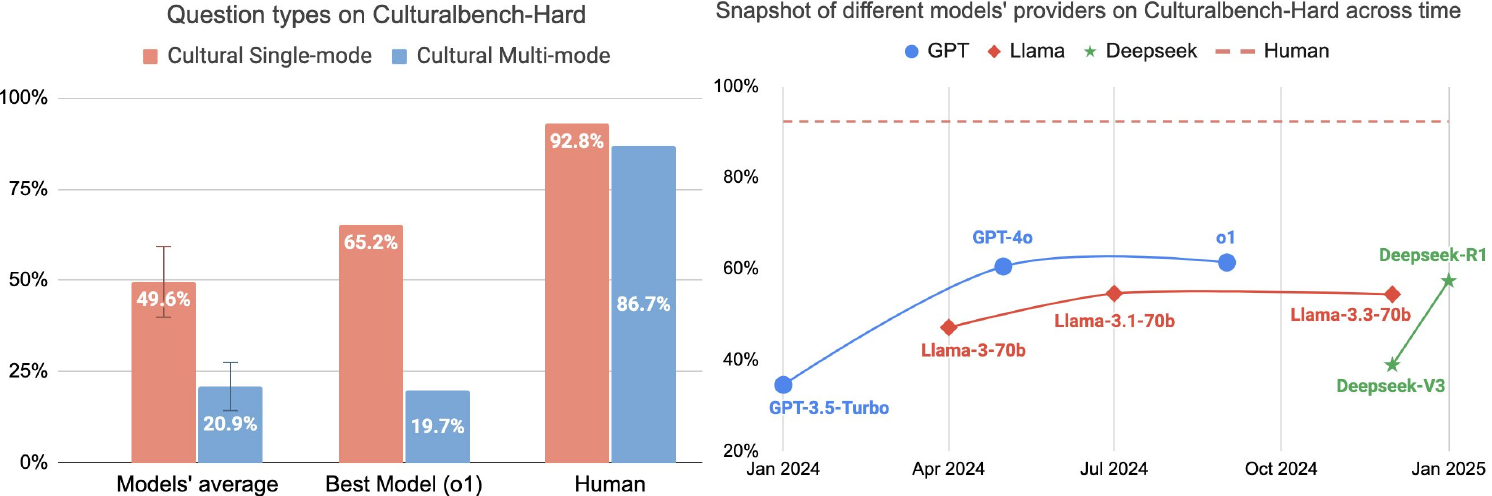}}
    \vspace{-2em}
    \caption{Analysis on question type (Left) and time version  (Right). For question type, we demonstrate models struggle at answering questions with multi-modes (more than one correct answers). For time version, we show the improved performance of models across time,  but some are approaching a performance ceiling.}
    \label{fig: question_type_time_version}
\end{figure*}

\paragraph{\benchmark-Hard.} We evaluate model performance based on the proportion of tasks in which the model can get all four options predicted correctly. For each task, an LM has to make four binary judgements (True/False) from the transformation of four options in each multiple choice question. To demonstrate robust cultural knowledge, we believe the LM has to accurately which option(s) are False as well as which option(s) are True. A random baseline can achieve $0.5^{4}$ = 6.25\%.

\subsection{Comparing LMs on two benchmarks across model family and size}

Due to space constraint, we show the performance of 23 representative models across model families and sizes on \benchmark-Hard in Fig. \ref{fig: benchmark_hard}. The corresponding Fig. for \benchmark-Easy is in Fig. \ref{fig: benchmark_easy}.

\paragraph{Models shows a huge performance gap with humans on \benchmark-Hard.} As shown in Fig. \ref{fig: benchmark_hard}, this setup is significantly more \textit{\textbf{challenging}} for current LMs, with accuracy ranging from 28.7\% for Cohere Aya-8b to 61.4\% for OpenAI o1. These scores are considerably lower compared to the human baseline of 92.4\%, highlighting the difficulty of the task even for the most advanced models.

\paragraph{Models performance improves as model size increases.} 
In Fig. \ref{fig: benchmark_hard}, we present the performance of models from six different families, such as GPT, Llama, and Qwen. 
Overall, the results demonstrate a trend of improved performance as model size increases. 
For example, within the OpenAI model family, the models show a clear progression in accuracy: GPT-3.5 Turbo achieves 34.5\%, GPT-4o attains 60.4\% and o1 attains 61.5\%. Similarly, in Fig \ref{fig: benchmark_hard}, in the Llama-3.1 family, 8B achieves 36.0\%, 70B has 54.6\%, and 405B has 51.93\%. The 70b model has a much higher score than the 8b model, albeit the inconsistent pattern in Llama-3.1 405B.  This pattern is consistent across most of the model families, indicating that larger models generally have better cultural knowledge.

\paragraph{All models tested have a huge performance difference between the two setups of \benchmark.} For instance, the best-performing model, o1, achieves 89.6\% accuracy on \benchmark-Easy but drops to 61.4\% accuracy on \benchmark-Hard, 
whereas the human baselines are both 92.4\% and the random baselines are 25\% and 6.25\% respectively.
We hypothesize that the models can guess the most possible answer on \benchmark-Easy under the multiple-choice setting. To investigate this hypothesis, we compute the embedding for the country name and separately for each option using OpenAI text-embedding-3-small. 

By using a simple heuristic of choosing the option with highest cosine similarity with the country name (e.g. Bangladesh), we attain 40.4\% accuracy. This is intriguing as it is substantially above the random baseline for multiple-choice setup in \benchmark-Easy (25\%), without needing considering the question at all. We find that the cosine similarity difference between the correct option and the country name is significantly higher than the difference between options average and the country (0.166 vs. 0.145; Kruskal-Wallis $p$-value$\leq$0.01). This shows the possibility of models guessing based on one (out of many possible) heuristics in multiple-choice setup without understanding (or even \textit{knowing}) the question. This stresses the importance on using the binary (True/False) for each of the four options per question in \benchmark-Hard to accurately assess cultural knowledge of LMs. 

\begin{figure*}[t]
    \centering
\resizebox{0.95\textwidth}{!}{%
\includegraphics[width=\textwidth]{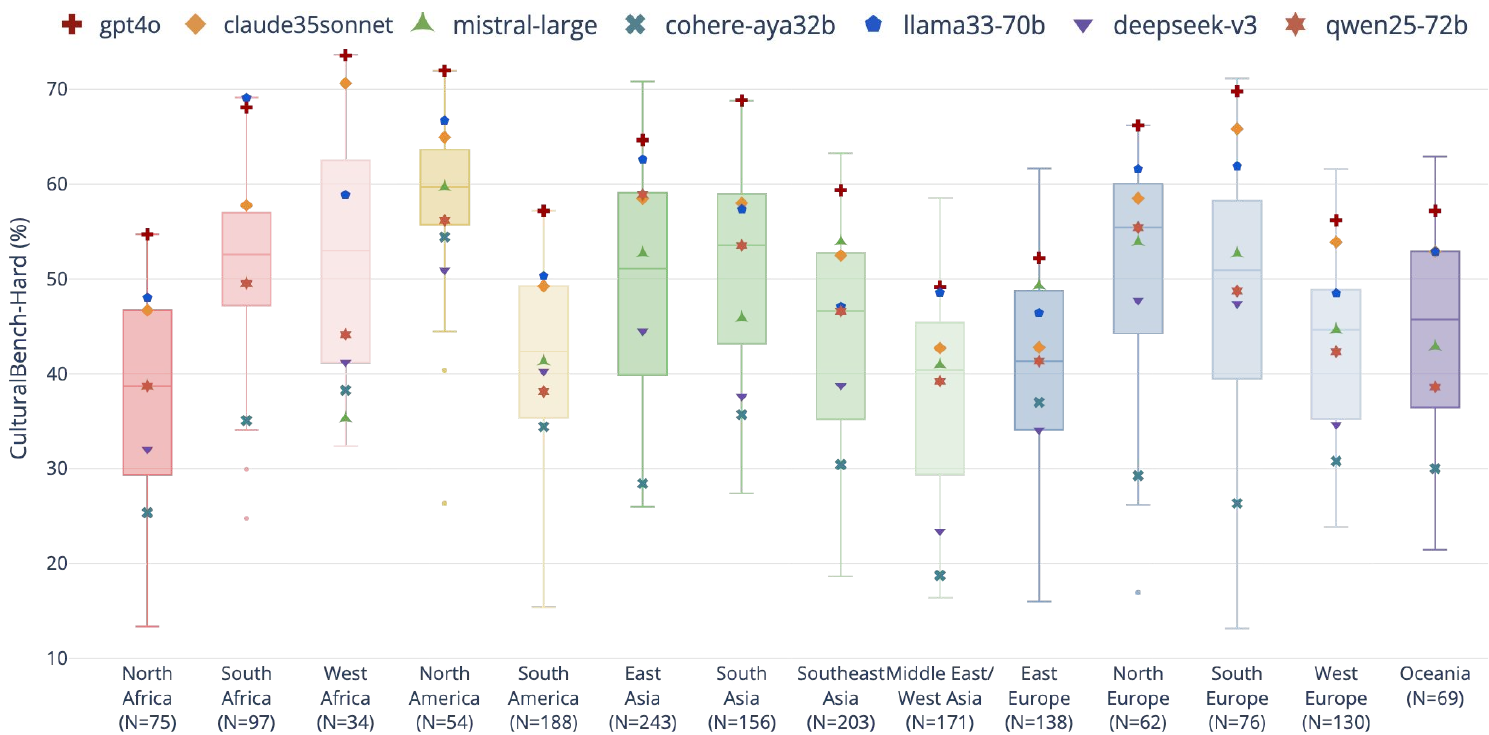}}
    \vspace{-0.87em}
    \caption{Models performance by different providers on various geographic regions. We further compare seven representative models (GPT-4o, Claude Sonnet, Mistral Large, Cohere Aya 32b, Llama-3.3-70b, DeepSeek V3 and Qwen-2.5 72b) from different model families. We avoid plotting OpenAI o series and DeepSeek R1 as they spend substantial reasoning tokens, which might be unfair comparison.}
    \label{fig: provider_region}
\end{figure*}

\subsection{Investigating effects of question type and time version of models}

\paragraph{LMs show distinct gaps between question types, unlike humans.} We evaluate the performance based on question types -- (1) Single-mode (N=1554) and (2) Multi-mode (N=142). The first type refers to the questions with only one correct answer while the second type includes questions with multiple correct answers, as explained in Section \ref{sec: modes_questions}. All the correct answers are majority-voted. In Fig. \ref{fig: question_type_time_version} (Left), the average across all models shown in Fig. \ref{fig: benchmark_hard} is 49.6\% on Single-mode questions and 20.9\% on Multi-mode questions, revealing a significant gap of 28.7\% between the two. Similarly, the best model (o1) exhibits a 45.5\% performance difference between these question types. In contrast, human baselines show only a 6.1\% difference, indicating that humans handle cultural nuances (where one question could have more than one correct answers) more effectively than models. This discrepancy suggests that models struggle to account for cultural nuances due to their mode-seeking tendencies, supporting a similar observation by \citet{tajwar2024preference}.

\paragraph{Model providers improve on cultural knowledge of models across time, but are approaching a performance ceiling.} In Fig. \ref{fig: question_type_time_version} (Right), we evaluate three popular models familiy (GPT, Llama, Deepseek) across released time. Overall, all model families demonstrate an increasing trend in performance across time. However, most model families show gradual improvements for their newer versions. For instance, the Llama family shows only a slight improvement from Llama-3.1-70b to Llama-3.3-70b, compared to the improvement between Llama-3-70b and Llama-3.1-70b.

\subsection{Studying different providers' LMs on questions from different regions}

We include detailed performance of models across different family and sizes (shown in Fig \ref{fig: provider_region}) to understand how well different models performance in questions relating to different geographic regions at a sub-continent level.

\paragraph{Models perform better in questions relating to North America, North Europe and South Asia.} From Fig. \ref{fig: provider_region}, it is evident that models achieve higher performance averages in regions like North America (57.9\%), North Europe (51.8\%) and South Asia (51.5\%).  We hypothesize that the higher performance in these regions can be attributed to several factors including their representation on web-data used for model training \citep{longpre2023pretrainersguidetrainingdata} and the proportion of annotators recruited from these regions by LM providers to curate post-training alignment data. For instance, many annotators are known to be recruited from India as they have good English ability and costs substantially less than their counterparts in the United States \citep{indiacrowd}.

\paragraph{Models score lower in questions relating to South America, East Europe, and the Middle East.} Models exhibit lower performances on average in regions like South America (41.5\%), East Europe (41.5\%), and Middle East/West Asia (37.8\%). These disparities possibly suggest insufficient representation of cultural knowledge from these regions in training data at various stages.

\paragraph{GPT-4o leads in most regions, followed by Llama-3.3 70b and Claude-3.5 Sonnet.} GPT-4o consistently ranks highest across most regions among all tested models. Llama-3.3 70b shows strength in regions where cultural knowledge is traditionally less represented, such as South Africa, while Claude-3.5 Sonnet performs particularly well in other regions e.g., West Europe, West Africa. 

\paragraph{Chinese and European Model Providers are not stronger in cultural knowledge relating in their respective regions.} Despite claims of specialization in local languages, Qwen-25-72b, Deepseek V3 and Mistral Large do not outperform other models in their respective regions in terms of cultural knowledge. For example, Qwen-2-72b and Deepseek V3 score 58.8\% and 61.3\% on East Asia, while GPT-4o achieves 64.6\%. Similarly, Mistral Large underperforms in West Europe (44.6\%) compared to GPT-4o (56.2\%). These results suggest that model providers based out of specific regions do not necessarily have advantages in cultural knowledge of their regions.

\section{Related Work}

\begin{table*}[ht!]
    \centering
    \resizebox{\textwidth}{!}{%
    \begin{tabular}{c|c|c|c|c|c|c|c|c}
        \toprule
        \textbf{Benchmark} & \textbf{Task} & \textbf{\makecell[c]{\# Annotators per\\Qn ($\uparrow$)}} & \textbf{\makecell[c]{Verified Qn Coverage\\(Verified \#/Total \#) ($\uparrow$)}} & \textbf{\makecell[c]{Data Filtering by\\Majority Votes}} & \textbf{\makecell[c]{Topic\\Inclusion}} & \textbf{\makecell[c]{\# Topic\\($\uparrow$)}} & \textbf{Source} & \textbf{\makecell[c]{Best Model Performance\\($\downarrow$)}} \\
        \midrule
        \makecell[c]{FORK \\\citep{palta-rudinger-2023-fork}} & \makecell[c]{Text-based short-form\\QA} & 2 & 100\% (184/184) & \ding{56} & \makecell[c]{Predefined\\set} & 1 & Human & \makecell[c]{74.7\% (Average for BERT\\series models)} \\
        \midrule
        \makecell[c]{BERTAQA \\\citep{etxaniz2024bertaqa}} & \makecell[c]{Text-based short-form\\QA} & NA & NA & NA & NA & 12 & Web & 91.7\% (GPT-4) \\
        \midrule
        \makecell[c]{CVQA \\ \citep{romero2024cvqa}} & Visual QA & \makecell[c]{2 (including\\author of Q)} & 100\% (10,374/10,374) & \ding{56} & \makecell[c]{Predefined\\set} & 10 & Humans & 75.4\% (GPT-4o) \\
        \midrule
        \makecell[c]{NormAd\\ \citep{Rao2024NORMADAB}} & \makecell[c]{Text-based short-form\\QA} & 2 & 18.5\% (480/2.6K) & \ding{56} & \makecell[c]{Predefined\\set} & 4 & Web + LLM & 87.6\% (GPT-4) \\
        \midrule
        \makecell[c]{Blend \\\citep{Myung2024BLEnDAB}} & \makecell[c]{Text-based short-\\form/long-form QA} & 5 & 0\% (0/500) & \ding{56} & \makecell[c]{Discovery-\\based} & 5 & \makecell[c]{Human +\\LLM} & 85.5\% (GPT-4) \\
        \midrule
        \makecell[c]{CULTURALBENCH\\ (Our Work)} & \makecell[c]{Text-based short-form\\QA} & 5 & 100\% (1696/1696) & \ding{52} & \makecell[c]{Discovery-\\based} & 17 & \makecell[c]{Human +\\LLM} & 61.4\% (o1) \\
        \bottomrule
    \end{tabular}
    }
    \caption{Comparison of existing cultural benchmarks on three criteria. Relative to existing benchmarks, \benchmark is \textit{robust}, \textit{diverse} and \textit{challenging}. Verified Qn Coverage refers to the human quality checks on the final collected questions on the benchmark, rather than intermediate steps of data collection. Best Model Performance refers to the average scores attained by best performing model on benchmark, with the model in parenthesis. Reported metrics are as follows: Candle: Precision; CultureAltas: F1; Normad: accuracy; Blend: accuracy; \benchmark: accuracy.}
    \label{table:compare_benchmarks}
\end{table*}

Multicultural knowledge evaluation of LMs has been widely investigated through building extensive knowledge bases \citep{shi2024culturebank, keleg2023dlama}; using socio-cultural surveys like World Value Survey \citep{durmus2023towards,tao2023auditing,ramezani2023knowledge}; and generating more training data \citep{Li2024CultureLLMIC}. Here, we select four representative benchmarks with comparable model evaluation results, highlighting their limitations and the gaps that our \benchmark aims to fill in Table \ref{table:compare_benchmarks}.

\paragraph{Insufficient Quality Verification.}  Existing cultural benchmarks usually conduct quality checks during the intermediate steps on data collection such as the relevance of web-scraped knowledge \citep{Fung2024MassivelyMK}, commonality of knowledge \citep{Nguyen2022ExtractingCC}. Blend asked humans to directly curate answers and aggregating those inputs to form questions but did not verify the final questions by humans \citep{Myung2024BLEnDAB}. Normad verified part of the rule-of-thumbs but with two humans only \citep{Rao2024NORMADAB}. As cultural knowledge is not easily verifiable for correctness, it is essential to have reliable annotations on the final set of questions (as given to LMs) by having expert human verification on the full set of questions and filtering out questions without consensus answers.

\paragraph{Predefined cultural topics.} Many benchmarks have topics predefined prior to data collection, meaning that they are unlikely to fully capture the multi-faceted natured of cultural knowledge \citep{adilazuarda2024towards}. Many prior works topics focus on narrow topics such as food \citep{Nguyen2022ExtractingCC}, dating \citep{Fung2024MassivelyMK}, social etiquette like dining \citep{palta2023fork,dwivedi2023eticor}, visiting \citep{Rao2024NORMADAB}, and special elements in wider society like religions \citep{Nguyen2022ExtractingCC}. While Blend uses a discovery-based approach \citep{Myung2024BLEnDAB}, \benchmark extends it to identify more \textit{\textbf{diverse}} topics (6 vs. 17) with details in Fig. \ref{fig: topic}.

\paragraph{Over-reliance on Web Sources.} Existing benchmarks often rely on web sources directly such as web corpus \citep{Nguyen2022ExtractingCC}, Wikipedia \citep{Naous2023HavingBA}, and incorporated with LMs' generation \citep{Rao2024NORMADAB,Fung2024MassivelyMK}. These non-human written benchmarks may not be \textit{\textbf{challenging}} since the scraped web sources may be used during models pretraining \citep{Petroni2019LanguageMA} and LM generations may inherit potential cultural biases \citep{Arora2022WhyEB,cao-etal-2023-assessing, liu2024best}. Given the highest performance of models range from 81.4\% to 93.1\% in the existing benchmarks by relatively weaker models (e.g. GPT-3/3.5/4) in Table \ref{table:compare_benchmarks}, those benchmarks are likely not sufficiently \textbf{\textit{challenging}} for modern frontier LMs (e.g. o1). Our proposed \benchmark is substantially more difficult with the best model (OpenAI o1) only reaching 61.5\%, far from the human performance of 92.4\%

\section{Conclusion}

Inspired by the human-AI red teaming, we establish \benchmark: a robust, diverse, and challenging benchmark for measuring LMs' culture knowledge. 
We hope the community can build upon our work to accelerate our progress in improving cultural knowledge of LMs.

\clearpage
\clearpage
\section*{Limitations}

While \benchmark has several advantages over existing cultural benchmarks, we would also like to clarify some of its current limitations as well as ways to address them in the future.

\paragraph{Multilingual vs. Multicultural.} We develop an English-only benchmark as the initial step in evaluating models' cultural knowledge. This approach facilitates fair comparisons of cultural understanding across different regions. For instance, in underrepresented regions such as Bangladesh, the availability of training data in local languages is often limited. As a result, models lacking sufficient exposure to these languages may struggle to comprehend questions phrased in them \citep{yong2023low}. By employing an English-only benchmark, we can assess models' cultural knowledge regarding these underrepresented areas without considering their (lack of) proficiency in low-resource languages. Additionally, prior research on multilingual models' emotional understanding \citep{havaldar2023multilingual} and reasoning skills \citep{liu2023multilingual} indicates that a model's multilingual capabilities may not necessarily correlate with its multicultural competencies.

\paragraph{Small sample of human verifiers on subjective cultural knowledge.} Due to the limitations of crowd-sourcing platforms like Prolific, the number of available annotators from underrepresented regions, such as Bangladesh, is quite small (fewer than 30 active human annotators). As a result, we were able to recruit only five annotators for consistency verification. To enhance the robustness of our dataset, we allow human verifiers to select multiple labels for each question, ensuring that all possible answers are captured. Additionally, we establish a strict majority-vote threshold (majority votes $\geq$ 4 out of 5). During the annotation process, we also provide two extra options: \textit{``I don't have knowledge''} and \textit{``This question is unanswerable''} -- to enable annotators to indicate when they cannot provide a response, as illustrated in Appendix Fig. \ref{fig:qc-guidance-step-1} and \ref{fig:qc-sample-question}.

\paragraph{Further fine-grained culture classification.}
\label{sec: further-finegrain-culture}
We noticed that the country/region classification adopted by our \benchmark may not capture the cultural diversity within each region. However, the data annotation platform we accessed does not have a further fine-grain classification when recruiting human annotators for most of the regions except for the United States and the United Kingdom. To capture the diversity on these two countries, we revisited the data that have been filtered by having not enough majority votes and with mostly responses of \textit{``I don't have knowledge''}. For example, questions asking for the Welsh custom in the United Kingdom may not be answerable for people living in England. Then, we conducted a second round of human quality check by assigning those questions for the specific groups of human annotators (e.g., people living in Wales in the United Kingdom), as explained in Section \ref{sec: human_qc}. We hope to see more data annotation tools for different local cultures to facilitate more fine-grained cultural data collection.

\section*{Risk} Our benchmark may unintentionally increase existing biases in LM Cultural Knowledge evaluation. Even through we tried our best to capture cultural questions from different regions, some cultural perspectives might be overrepresented or underrepresented (due to the availability/perspectives of annotators on the Prolific platform), leading to skewed evaluations.

\section*{Ethical Considerations} 
Our data collection has been reviewed by the university's IRB board to ensure it has no harm on human annotators. We pay annotators according to our vendor (Prolific)'s guidance, which is higher than the local wage requirement. Before annotation, we explain the annotation task, and how their data will be collected and transformed into the quiz questions for models. With such understanding, they consented to their annotations being used in such a manner. Our annotation guidance has specifically asked annotators to not include their personal identifiable information when giving their responses. Before human verification, our internal team has reviewed the collected data to ensure there is no harmful or unsafe context such as sexual or violence content.

\subsubsection*{Acknowledgments}
We thank the users of our \method platform for their helpful feedback and comments. This research was supported in part by DARPA under the ITM program (FA8650-23-C-7316).

\clearpage

\bibliography{custom}

\appendix

\clearpage

\begin{figure*}[ht]
    \centering
\resizebox{\textwidth}{!}{%
\includegraphics[width=\textwidth]{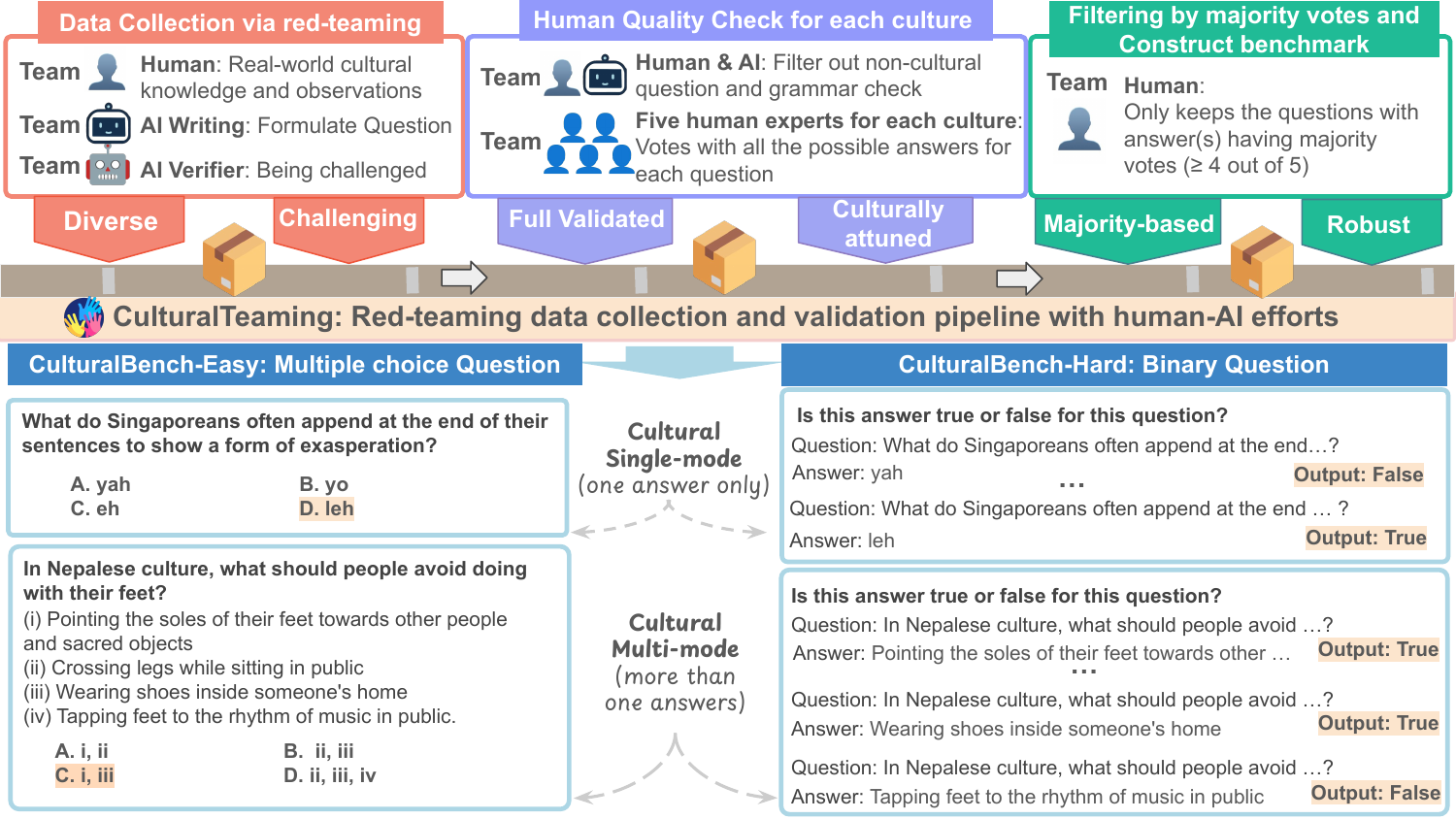}}
    \vspace{-2em}
    \caption{Overview of human-AI collaborating red-teaming data collection and validation to construct \benchmark.}
    \label{fig:Concept_and_data}
\end{figure*}
\clearpage

\section{Supplementary detail and experiments}

\textbf{Criteria on selecting underrepresented region/culture.} Our goal in CulturalBench is to cover diverse regions/cultures that are typically neglected in LLM evaluation and training due to factors such as the dominance of English-based internet data and researcher geographical distribution. We consider "underrepresented" regions as those with significant populations but have disproportionately low representation in LLM development. In practice, beyond English-speaking regions (primarily the United States) and mainland China, many regions qualify as "underrepresented." While we didn't apply strict criteria for defining "underrepresented" regions, we used heuristics including country/region populations and numbers of native language speakers to build our initial set of regions. We then filtered for regions with at least 10 active annotators on Prolific (our annotation platform vendor) as our validation protocol requires a minimum of 5 annotators from each region.

\textbf{Additional Multilingual Experiment.} We conducted an additional experiment with translated questions. We collaborated with 3 native volunteers to translate questions related to China (Simplified Chinese), Taiwan (Traditional Chinese), and Hong Kong (Cantonese). We then evaluated two models: GPT-4o and Qwen 2.5 72B (from a Chinese LLM provider). The results are summarized below:

\begin{table}[ht]
\centering
\small
\resizebox{\columnwidth}{!}{%
\begin{tabular}{lccc}
\toprule
\textbf{Model} & \textbf{Full} & \textbf{Single-Mode} & \textbf{Multi-Mode} \\
\midrule
GPT-4o (English subset) & 90.6\% & 91.3\% & 80.0\% \\
GPT-4o (Chinese) & 87.9\% & 88.4\% & 80.0\% \\
Qwen 2.5 72B (English subset) & 85.9\% & 85.5\% & 90.0\% \\
Qwen 2.5 72B (Chinese) & 79.9\% & 79.7\% & 80.0\% \\
\bottomrule
\end{tabular}}
\caption{Easy Version (Accuracy \%): Performance comparison across subsets and mode types.}
\label{tab:multilingual_easy}
\end{table}

\begin{table}[ht]
\centering
\small
\resizebox{\columnwidth}{!}{%
\begin{tabular}{lccc}
\toprule
\textbf{Model} & \textbf{Full} & \textbf{Single-Mode} & \textbf{Multi-Mode} \\
\midrule
GPT-4o (English subset) & 62.4\% & 63.3\% & 50.0\% \\
GPT-4o (Chinese) & 57.0\% & 57.6\% & 50.0\% \\
Qwen 2.5 72B (English subset) & 55.7\% & 54.7\% & 70.0\% \\
Qwen 2.5 72B (Chinese) & 52.3\% & 52.5\% & 50.0\% \\
\bottomrule
\end{tabular}}
\caption{Hard Version (Accuracy \%): Performance comparison across subsets and mode types.}
\label{tab:multilingual_hard}
\end{table}

Across both models, performance on translated questions is slightly lower than performance on the English versions, though the differences are not dramatic. This suggests that our English-only approach provides a reasonable assessment of cultural knowledge while avoiding potential confounds from language proficiency variations.

\textbf{Another way of evaluation: Distribution-based Evaluation} We explored using GPT-4o to calculate probability distributions over answers, computing linear probability using log probabilities. Our analysis of these linear probabilities revealed a highly right-skewed distribution: In total, there are 6784 instances (binary statement questions) for 1696 questions.

\begin{itemize}
    \item (50-60\%]: 1.7\% of responses
    \item (60-70\%]: 1.8\% of responses
    \item (70-80\%]: 2.0\% of responses
    \item (80-90\%]: 2.7\% of responses
    \item (90-100\%]: 26.4\% of responses
    \item Exactly 100\%: 65.4\% of responses
\end{itemize}

This highly right-skewed distribution reveals that GPT-4o is typically extremely confident in its answers (>90\% confidence in 91.78\% of cases). This overconfidence suggests that using probability distributions as an evaluation metric may not meaningfully change the conclusions, as models rarely express uncertainty even when incorrect.

While we appreciate the reviewer's suggestion about rephrasing questions to provide more context, we believe the current format effectively reveals important limitations in models' cultural understanding - specifically their tendency to provide singular, overly confident answers in scenarios where cultural nuance requires acknowledging multiple valid perspectives.

\begin{table}[ht]

\resizebox{0.5\textwidth}{!}{%
\centering
\begin{tabular}{p{0.31\textwidth}p{0.25\textwidth}p{0.22\textwidth}}
\toprule
\textbf{Paper} & \textbf{Task} & \textbf{Will it be listed in which table for the revised version?} \\
\midrule
CultureBank: An Online Community-Driven Knowledge Base Towards Culturally Aware Language Technologies \cite{shi2024culturebank} & Knowledge base + automatic evaluation & Not specifically for QA task - Extensive table \\
\midrule
Auditing and mitigating cultural bias in llms//Cultural Bias and Cultural Alignment of Large Language Models  & Using WVS to evaluate on different cultural prompting variations \cite{tao2023auditing}  & Not QA eval - NA \\
\midrule
DLAMA: A Framework for Curating Culturally Diverse Facts for Probing the Knowledge of Pretrained Language Models \cite{keleg2023dlama} & Knowledge base + Probing  & Not specifically for QA task - Extensive table \\
\midrule
Towards Measuring the Representation of Subjective Global Opinions in Language Models \cite{durmus2023towards} & Use WVS/PEW to measure representation (percentage) of people's opinions.  & Not specifically for QA task - Extensive table \\
\midrule
Knowledge of cultural moral norms in large language models \cite{ramezani2023knowledge}& Probe LMs to get cultural moral norms  & Not QA eval - NA \\
\midrule
Towards measuring and modeling" culture" in llms: A survey \cite{adilazuarda2024towards}  & Not QA eval - NA \\
\midrule
CultureLLM: Incorporating Cultural Differences into Large Language Models \cite{Li2024CultureLLMIC} & Prompting the LLMs to reduce cultural bias.  & Not QA eval - NA \\
\midrule
Massively multi-cultural knowledge acquisition \cite{Fung2024MassivelyMK}& lm benchmarking \& Building knowledge base of cultural knowledge + evaluation & Not specifically for QA task - Extensive table \\
\midrule
(Candle:) Extracting Cultural Commonsense Knowledge at Scale \cite{Nguyen2022ExtractingCC} & Building knowledge base + evaluation & Not specifically for QA task - Extensive table \\
\midrule
BLEND: A Benchmark for LLMs on Everyday Knowledge in Diverse Cultures and Languages \cite{Myung2024BLEnDAB} & Text-based short-form/long-form QA & Continue to keep in Table 1 \\
\midrule
NormAd: A benchmark for measuring the cultural adaptability of large language models \cite{Rao2024NORMADAB} & Text-based short-form QA  & Continue to keep in Table 1 \\
\midrule
FORK: A Bite-Sized Test Set for Probing Culinary Cultural Biases in Commonsense Reasoning Models \cite{palta2023fork} & Text-based short-form QA  & Newly added to Table 1 \\
\midrule
Eticor: Corpus for analyzing llms for 747 etiquettes. \cite{dwivedi2023eticor}& Text-based subjective preference/judgement eval & Not specific to QA - Extensive table \\
\midrule
Having beer after prayer? measuring cultural bias in large language models. \cite{Naous2023HavingBA}& Analysis model response/ wvs  & Not QA eval - NA \\
\midrule
CVQA: Culturally-diverse Multilingual Visual Question Answering Benchmark \cite{romero2024cvqa} & Visual QA  & Newly added into Table 1 \\
\midrule
CulturalVQA Benchmarking Vision Language Models for Cultural Understanding \cite{nayak-etal-2024-benchmarking} & Visual open-end QA from Candle & Involving automatic eval and less relevant to text-based QA - Extensive table \\
\midrule
CaLMQA: Exploring culturally specific long-form question answering across 23 languages \cite{arora2024calmqaexploringculturallyspecific} & Text-based long-form QA + automatic eval  & Automatic eval - Extensive table \\
\midrule
BERTAQA: How Much Do Language Models Know About Local Culture?\cite{etxaniz2024bertaqa} & Text-based short-form trivia QA & Newly added into Table 1 \\
\midrule
\benchmark (Our Work) & Text-based short-form QA & Table 1  \\
\bottomrule
\end{tabular}}
\caption{(Expanded table) Related Work on Cultural Awareness in Language Models}
\label{table:compare_benchmarks_extension}
\end{table}

\textbf{Extension table for related work.} In Table \ref{table:compare_benchmarks}, we focus specifically on QA benchmarks. Here, we incorporated additional relevant wor to provide a more comprehensive comparison in Table \ref{table:compare_benchmarks_extension}.

\clearpage

\begin{figure*}[ht!]
    \centering
\resizebox{\textwidth}{!}{%
\includegraphics[width=\textwidth]{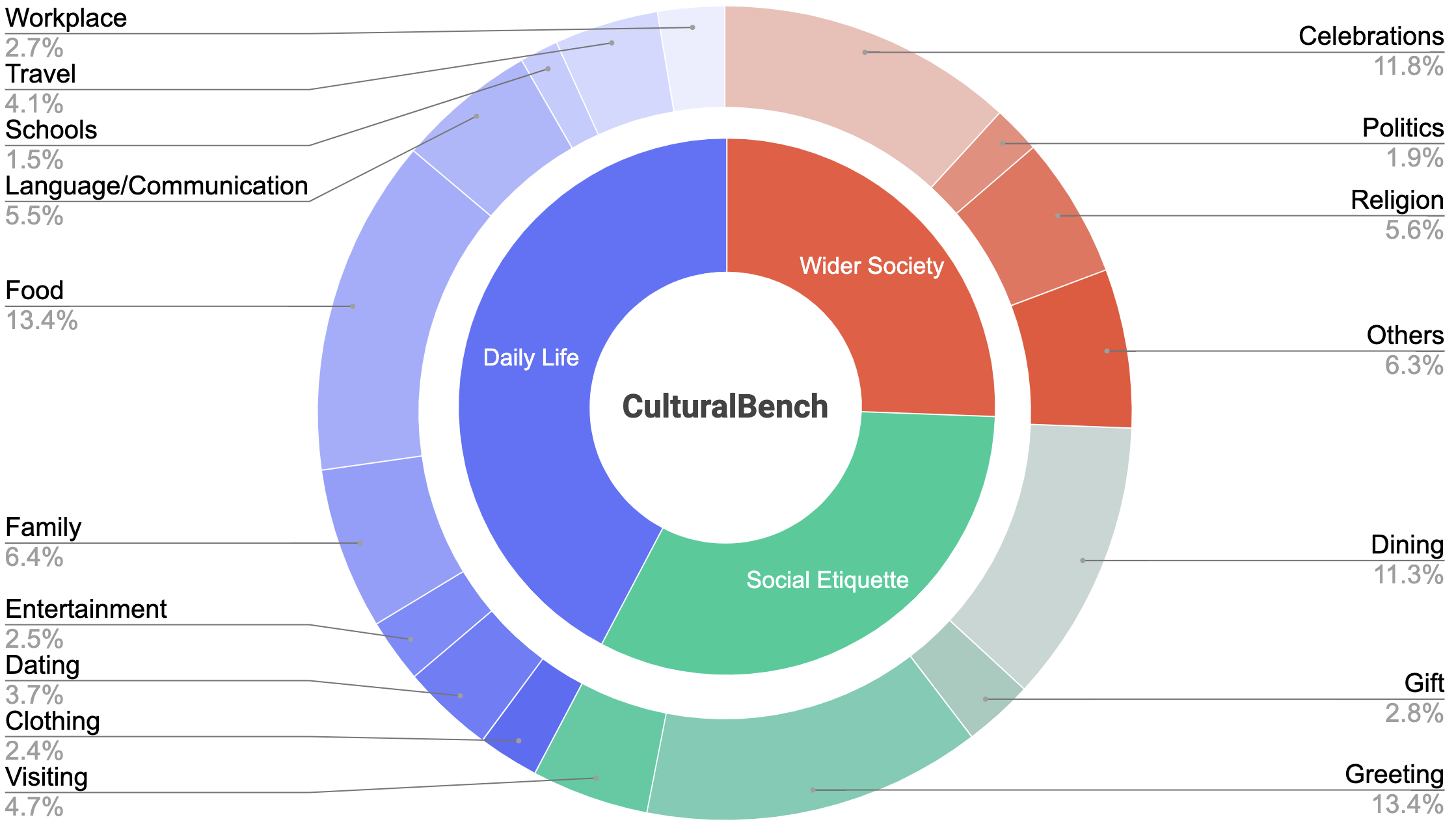}}
    \vspace{-2em}
    \caption{\benchmark covers 17 \textit{\textbf{diverse}} cultural topics organized into three overarching categories.}
    \label{fig: topic}
\end{figure*}
\clearpage
\section{Definition, examples and prompts for cultural topics}
\label{app: definition of topics}

\begin{itemize}
        \item Daily life: universal concepts among cultures
            \begin{enumerate}
                \item Clothing: Fashion, Jewelry trend;
                \item Food: cuisine, drinks;
                \item Entertainment e.g. game, movie, music, sports;
                \item Language/Communication e.g. linguistic, languages, general ways of communication that are not related to greeting people; 
                \item Schools e.g. school setting, students, education;
                \item Workplace e.g. business setting, employee, employer; 
                \item  Travel/transport: ways of transport; 
                \item Dating/marriage: committed relationship, romantic relationship; 
                \item  Family: children, parents, extended family relationship;
            \end{enumerate}
        \item Social Etiquette 
            \begin{enumerate}
                \item Greeting: what ways of greeting, gesture, how to greet people e.g. opposite sex, elderly;
                \item Dining: social rules related to dining e.g. which hands, how to pay; 
                \item Gift: social rules about gift sending and receiving;
                \item  Visiting and punctuality: rules about the concept of time, attending meetings, visiting friends; 
            \end{enumerate}
        \item Wider society: specific concepts for certain cultures
            \begin{enumerate}
                \item Celebrations e.g., celebrations, festival, holidays; 
                \item  Politics e.g. laws, rules by organization;
                \item Religion e.g. religious rituals, religious holidays;
                \item Others: any other topics or something in general.
            \end{enumerate}
\end{itemize}

\paragraph{Prompt for classification:}
\texttt{
What is the most relevant topic for this question? Topics: <topic description and examples above> Instruction: Choose only one most relevant topic among the 17 topics. Only output the topic name e.g. Clothing, Food. Question: <Question>
}
\clearpage
\label{app: example_per_topic}
\begin{table*}
    \centering
    \resizebox{\textwidth}{!}{
    \begin{tabular}{c|c}
    \toprule
         \textbf{Topic}  &\textbf{Example of question} \\
         \midrule
         \multicolumn{2}{c}{\textbf{Daily Life}}\\
         \midrule
         Clothing &  \makecell[l]{What type of attire do Bengali males traditionally wear on Eid day?\\A. Kurta Pajama\\B. Suit and Tie\\C. T-shirt and Jeans\\D. Player's team apparel and gear}\\
         \midrule
         Dating & \makecell[l]{At what age do people commonly get married in German culture?\\A. In their mid-30s\\B. Right after high school\\C. In their early 20s\\D. Teenage years.}\\
         \midrule
         Entertainment & \makecell[l]{What is a popular activity among South Africans to relax during both summer and winter?\\A. Going for a braai\\B. Hosting a backyard barbecue\\C. Heading to a beach bonfire\\D. Camping}\\
         \midrule
         Family & \makecell[l]{What is not a common way for people to show respect to their grandparents in Moroccan culture?\\A. Kiss their grandparents' hands\\B. Kiss their grandparents on the forehead\\C. Bow to their grandparents\\D. Hug their grandparents}\\
         \midrule
         Food & \makecell[l]{In Indian culture, what is a common way of eating meals among locals?\\ Read the following statements and\\select the option that include all the appropriate statements for this question.\\
         (i) Using chopsticks\\(ii) Eating with a fork and knife\\(iii) Eating using hands/fingers\\(iv) Eating with a spoon only.\\A. ii, iv\\B. iii, iv\\C. i, iv\\D. i, iii, iv}\\
         \midrule
         \makecell[l]{Language/\\Communication} & \makecell[l]{What do Singaporeans often append at the end of their sentences as a linguistic quirk\\unique to their culture to show a form of exasperation?\\A. "yah"\\B. "lah"\\C. "eh"\\D. "yo"}\\
         \midrule
         Travel & \makecell[l]{In Indonesian culture, how do people mostly spend their time on the road?\\A. Bicycles\\B. Motorcycles\\C. Cars\\D. Scooters}\\

         \midrule
         Workplace & \makecell[l]{In Taiwanese culture, what is the most common question people will ask about your work?\\A. Do you like your job?\\B. How much do you make?\\C. What are your job benefits?\\D. Can you work remotely?}\\ 
        \bottomrule
    \end{tabular}
    }
    \caption{Examples of \benchmark for each topic. (Part 1)}
    \label{tab:topics_model_examples}
\end{table*}

\clearpage
\begin{table*}
    \centering
    \resizebox{\textwidth}{!}{
    \begin{tabular}{c|c}
    \toprule
         \textbf{Topic}  &\textbf{Example of question} \\
         \midrule
         \multicolumn{2}{c}{\textbf{Social Etiquette}}\\
         \midrule
         Dining & \makecell[l]{In Hong Kong culture, how do you signal to the server that you would like more water\\added to your tea while dining in local Chinese restaurants?\\A. Remove the lid of the teapot\\B. raise the cup and show to the server\\C. Leave the teapot empty on the table\\D. Point at the teapot}\\
         \midrule
         Greeting & \makecell[l]{In Yoruba culture of Nigera, how do young people traditionally greet elders?\\A. Shake hands while maintaining eye contact.\\B. Prostrate as a male and go on your knees as a female.\\C. Bow slightly without making any physical contact.\\D. Give a brief nod and smile.}\\
         \midrule
         Gift & \makecell[l]{In Russian culture, there are many gifts which are avoided because they represent bad omens\\or bad luck. Which of the following items is a very bad gift for your Russian friend?\\A. Even numbers of flowers less than a dozen\\B. Artwork\\C. Chocolate chip cookies\\D. Baseball caps}\\
         \midrule
         Visiting & \makecell[l]{In Peru, what is the cultural expectation regarding arrival time for a well\\planned birthday celebration event?\\A. Arriving just in time\\B. Arriving 15-30 minutes late\\C. Arriving an hour early\\D. Arriving 5-10 minutes early}\\
         \midrule
         \multicolumn{2}{c}{\textbf{Wider Society}}\\
         \midrule
         Celebrations & \makecell[l]{What do Vietnamese grandparents usually gift their grandchildren for the traditional\\Lunar New Year celebration?\\A. Cards with best wishes\\B. Traditional Foods and Snacks\\C. Monetary gifts\\D. Educational Materials}\\
         \midrule
         Religion & \makecell[l]{In Pakistani culture, what is the custom for Muslims regarding prayers on a specific day of the week?\\A. Praying at mosque on Sunday\\B. Offering Friday prayer\\C. Praying before lunch time\\D. Meditating on Friday morning.}\\
       \midrule
         Politics & \makecell[l]{In South Korea, only men are required to join the military.\\What are the alternative civic duties that can be performed instead of military service?\\A. Enrollment in educational programs for two years.\\B. Volunteering in community services for a year.\\C. Taking internship.\\D. None of the options}\\ 
       \midrule
         Others & \makecell[l]{How many seasons are traditionally recognized in Bangladeshi culture?\\A. 6 seasons\\B. 4 seasons\\C. 2 seasons\\D. 5 seasons}\\
        \bottomrule
    \end{tabular}
    }
    \caption{Examples of \benchmark for each topic. (Part 2)}
\end{table*}

\clearpage

\begin{figure*}[h]
    \centering
\resizebox{\textwidth}{!}{%
\includegraphics[width=\textwidth]{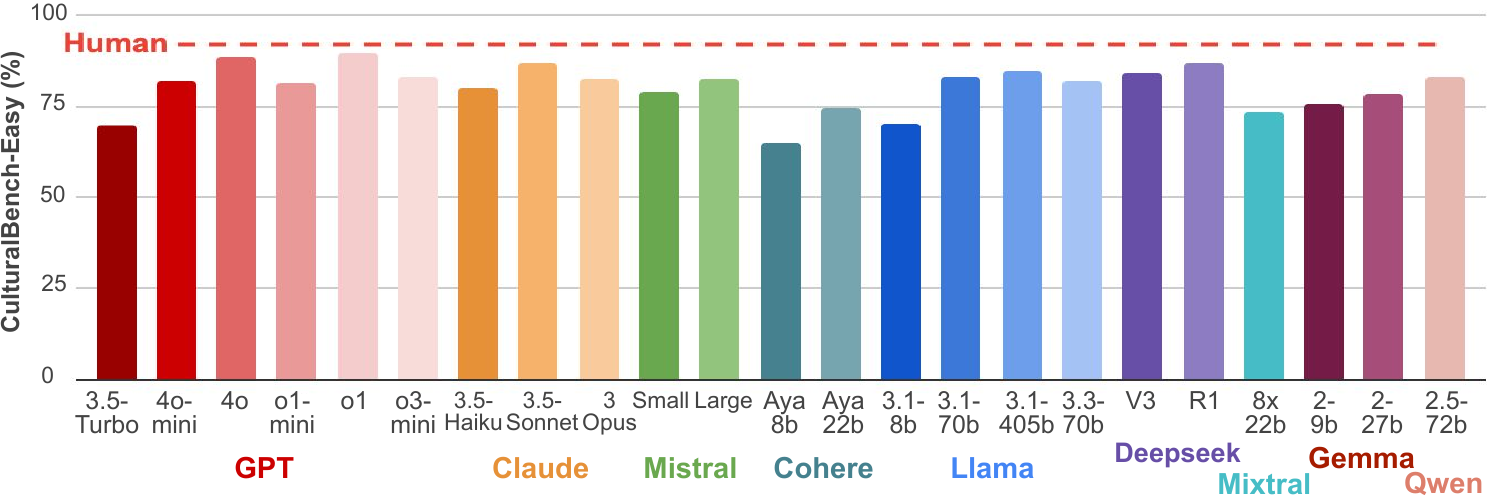}}
    \vspace{-2em}
    \caption{Models performance on \benchmark-Easy with random baseline at 25\% and human performance at 92.4\%}
    \label{fig: benchmark_easy}
\end{figure*}

\clearpage
\section{\benchmark statistics}
\label{app: country_distribution}
\begin{table}[h]
      \centering
        \begin{tabular}{c|c}
        \toprule
            \textbf{Country} & \textbf{Counts}\\ 
\midrule
\multicolumn{2}{c}{\textbf{North Africa} ($N = 75$)}\\ 
\midrule
Morocco & 38\\ 
Egypt & 37\\ 
\midrule
\multicolumn{2}{c}{\textbf{South Africa} ($N = 97$)}\\ 
\midrule
Zimbabwe & 39\\ 
South Africa & 58\\ 
\midrule
\multicolumn{2}{c}{\textbf{West Africa} ($N = 34$)}\\ 
\midrule
Nigeria & 34\\ 
\midrule
\multicolumn{2}{c}{\textbf{North America} ($N = 54$)}\\ 
\midrule
United States & 35\\ 
Canada & 19\\ 
\midrule
\multicolumn{2}{c}{\textbf{South America} ($N = 188$)}\\ 
\midrule
Chile & 35\\ 
Mexico & 49\\ 
Brazil & 33\\ 
Peru & 36\\ 
Argentina & 35\\ 
\midrule
\multicolumn{2}{c}{\textbf{East Asia} ($N = 243$)}\\ 
\midrule
South Korea & 41\\ 
Japan & 53\\ 
Taiwan & 54\\ 
Hong Kong & 36\\ 
China & 59\\ 
\midrule
\multicolumn{2}{c}{\textbf{South Asia} ($N = 156$)}\\ 
\midrule
India & 46\\ 
Bangladesh & 45\\ 
Nepal & 33\\ 
Pakistan & 32\\ 
\midrule
            \bottomrule
        \end{tabular}
    \caption{Distribution of questions across 45 countries in \benchmark (part 1)}
    \label{table:country_dist_1}
\end{table}
\newpage
\newpage
\begin{table}[h]
 \centering
        \begin{tabular}{c|c}
        \toprule
            \textbf{Country} & \textbf{Counts}\\ 
            \midrule 

\multicolumn{2}{c}{\textbf{Southeast Asia} ($N = 203$)}\\ 
\midrule
Vietnam & 33\\ 
Malaysia & 35\\ 
Philippines & 44\\ 
Indonesia & 32\\ 
Singapore & 32\\ 
Thailand & 27\\ 
\midrule
\multicolumn{2}{c}{\textbf{Middle East/West Asia} ($N = 171$)}\\ 
\midrule
Iran & 37\\ 
Israel & 42\\ 
Lebanon & 37\\ 
Saudi Arabia & 17\\ 
Turkey & 38\\ 
\midrule
\multicolumn{2}{c}{\textbf{East Europe} ($N = 138$)}\\ 
\midrule
Ukraine & 34\\ 
Czech Republic & 32\\ 
Romania & 38\\ 
Poland & 34\\ 
\midrule
\multicolumn{2}{c}{\textbf{North Europe} ($N = 62$)}\\ 
\midrule
United Kingdom & 33\\ 
Russia & 29\\ 
\midrule
\multicolumn{2}{c}{\textbf{South Europe} ($N = 76$)}\\ 
\midrule
Spain & 40\\ 
Italy & 36\\ 
\midrule
\multicolumn{2}{c}{\textbf{West Europe} ($N = 130$)}\\ 
\midrule
Netherlands & 51\\ 
France & 47\\ 
Germany & 32\\ 
\midrule
\multicolumn{2}{c}{\textbf{Oceania} ($N = 69$)}\\ 
\midrule
New Zealand & 32\\ 
Australia & 37\\ 
\midrule
            \bottomrule
        \end{tabular}
    \caption{Distribution of questions across 45 countries in \benchmark (part 2)}
    \label{table:country_dist_2}
\end{table}

\clearpage

\begin{table*}[ht]
   \centering
   \resizebox{1.1\textwidth}{!}{
    \begin{tabular}{c|c|c|c|c|c|c|c|c|c|c|c|c|c|c}
            \toprule
\textbf{\makecell[c]{}} & \textbf{\makecell[c]{North\\Africa}} &  \textbf{\makecell[c]{South\\Africa}} &  \textbf{\makecell[c]{West\\Africa}} &  \textbf{\makecell[c]{North\\America}} &  \textbf{\makecell[c]{South\\America}} &  \textbf{\makecell[c]{East\\Asia}} &  \textbf{\makecell[c]{South\\Asia}} &  \textbf{\makecell[c]{Southeast\\Asia}} &  \textbf{\makecell[c]{Middle East/\\West Asia}} &  \textbf{\makecell[c]{East\\Europe}} &  \textbf{\makecell[c]{North\\Europe}} &  \textbf{\makecell[c]{South\\Europe}} &  \textbf{\makecell[c]{West\\Europe}} &  \textbf{\makecell[c]{Oceania}} \\
\midrule 
count (\%) & 5.17 & 5.47 & 2.34 & 9.28 & 11.43 & 5.32 & 9.06 & 12.49 & 9.55 & 9.28 & 2.49 & 3.06 & 7.77 & 7.28\\
\midrule
\multicolumn{15}{c}{Ethnicity(\%)}\\ 
\midrule 
Asian & 0.0 & 2.07 & 0.0 & 19.92 & 0.33 & 98.58 & 91.67 & 94.86 & 7.91 & 0.41 & 60.61 & 4.94 & 7.28 & 21.76 \\ 
Mixed & 29.2 & 2.76 & 4.84 & 6.91 & 31.35 & 0.71 & 7.08 & 2.72 & 13.44 & 1.22 & 0.0 & 2.47 & 6.8 & 15.54 \\ 
Other & 27.01 & 0.0 & 0.0 & 7.32 & 22.77 & 0.0 & 0.83 & 2.11 & 33.99 & 0.81 & 3.03 & 2.47 & 0.49 & 5.18 \\ 
White & 42.34 & 10.34 & 0.0 & 42.28 & 44.88 & 0.0 & 0.42 & 0.3 & 44.27 & 97.56 & 34.85 & 88.89 & 83.01 & 57.51 \\ 
\midrule 
\multicolumn{15}{c}{Age group(\%)}\\ 
\midrule 
18-29 & 68.61 & 40.0 & 46.77 & 26.83 & 41.91 & 47.52 & 50.42 & 56.8 & 52.96 & 53.25 & 15.15 & 40.74 & 50.97 & 31.61 \\ 
30-39 & 29.93 & 33.1 & 46.77 & 27.64 & 39.93 & 33.33 & 41.67 & 25.98 & 33.99 & 31.71 & 28.79 & 24.69 & 31.07 & 28.5 \\ 
40-49 & 0.73 & 18.62 & 4.84 & 26.83 & 13.86 & 13.48 & 6.67 & 14.8 & 6.72 & 10.98 & 25.76 & 25.93 & 12.14 & 24.87 \\ 
50-59 & 0.0 & 6.9 & 1.61 & 13.01 & 2.97 & 5.67 & 1.25 & 2.42 & 2.77 & 4.07 & 22.73 & 6.17 & 2.43 & 7.25 \\ 
60 above & 0.73 & 1.38 & 0.0 & 5.69 & 1.32 & 0.0 & 0.0 & 0.0 & 3.56 & 0.0 & 7.58 & 2.47 & 3.4 & 7.77 \\ 
\midrule 
\multicolumn{15}{c}{Sex(\%)}\\ 
\midrule 
Female & 30.66 & 75.86 & 35.48 & 50.0 & 40.59 & 68.79 & 40.83 & 53.78 & 46.25 & 52.03 & 53.03 & 43.21 & 29.61 & 50.78 \\ 
Male & 67.15 & 24.14 & 64.52 & 49.59 & 59.41 & 29.79 & 58.75 & 45.32 & 52.96 & 47.56 & 46.97 & 56.79 & 69.9 & 48.19 \\ 
Prefer not to say & 2.19 & 0.0 & 0.0 & 0.41 & 0.0 & 1.42 & 0.42 & 0.91 & 0.79 & 0.41 & 0.0 & 0.0 & 0.49 & 1.04 \\ 
\midrule 
\multicolumn{15}{c}{Student Status(\%)}\\ 
\midrule
No & 35.77 & 53.79 & 53.23 & 79.67 & 62.05 & 58.16 & 44.58 & 60.42 & 44.27 & 56.91 & 84.85 & 51.85 & 66.5 & 63.73 \\ 
Yes & 52.55 & 41.38 & 41.94 & 13.01 & 31.02 & 29.79 & 45.83 & 29.91 & 45.85 & 31.71 & 9.09 & 46.91 & 27.18 & 21.76 \\ 
        \bottomrule
    \end{tabular}}
    \caption{Annotators demographic in Prolific for the whole dataset before filtering. We set two main recruitment criteria to ensure the recruited annotators have a deep understanding of culture of the targeted country or region: (1) Nationality. (2) Primary residence before age 18. See details in Section \ref{sec: human_qc}.}
\label{table:evaluation_models_on_question_type}
\end{table*}

\clearpage
\section{Zero-shot evaluation prompts}
\label{app: zero-shot-eval-prompts}
\textbf{\benchmark-Easy}

 Our evaluation is to ask the model in multiple choice setting. The zero-shot prompt is as follows to ensure the model only outputs one label (A, B, C, or D). We set the output token to be 1 for most of the models except for the reasoning models (e.g. o1, o1-mini, o3-mini and Deepseek r1) since they are not allowed.

\texttt{To answer the following multiple-choice question, you should choose one option only among A,B,C,D. Instruction: You must select one option among A,B,C,D. Do not output any other things.}

\texttt{Question:} \texttt{\textless Question\textgreater}

\texttt{A.} \texttt{\textless Option A\textgreater}

\texttt{B.} \texttt{\textless Option B\textgreater}

\texttt{C.} \texttt{\textless Option C\textgreater}

\texttt{D.} \texttt{\textless Option D\textgreater}

For multi-mode question, we included the instruction \textit{``Select the options with all applicable statements''} to ensure models considering all statements provided.

\textbf{\benchmark-Hard}

Our evaluation is to ask the model in binary setting (True/False). Our prompt is as follow to ensure the model only output one label (True/False). We set the output token to be 2 for most of the models except for the reasoning models (e.g. o1, o1-mini, o3-mini and deepseek r1) since they are not allowed.

\texttt{Question:} \texttt{\textless Question\textgreater}

\texttt{Answer:} \texttt{\textless Answer\textgreater}

\texttt{Is this answer true or false for this question? You must choose either True or False.'}

\clearpage
\begin{table*}[ht]
\resizebox{1.1\textwidth}{!}{
   \centering
    \begin{tabular}{c|c|c|c|c|c|c|c|c|c|c|c|c|c|c|c}
            \toprule
\textbf{\makecell[c]{Model}} & \textbf{\makecell[c]{Overall}}  & \textbf{\makecell[c]{North\\Africa}} &  \textbf{\makecell[c]{South\\Africa}} &  \textbf{\makecell[c]{West\\Africa}} &  \textbf{\makecell[c]{North\\America}} &  \textbf{\makecell[c]{South\\America}} &  \textbf{\makecell[c]{East\\Asia}} &  \textbf{\makecell[c]{South\\Asia}} &  \textbf{\makecell[c]{Southeast\\Asia}} &  \textbf{\makecell[c]{Middle East/\\West Asia}} &  \textbf{\makecell[c]{East\\Europe}} &  \textbf{\makecell[c]{North\\Europe}} &  \textbf{\makecell[c]{South\\Europe}} &  \textbf{\makecell[c]{West\\Europe}} &  \textbf{\makecell[c]{Oceania}} \\
\midrule
claude-3.5-haiku &80.07 & 80.0 & 81.44 & 88.24 & 80.7 & 76.19 & 81.07 & 81.53 & 85.78 & 72.51 & 76.81 & 84.62 & 84.21 & 77.69 & 81.43 \\
claude-3.5-sonnet &87.28 & 88.0 & 89.69 & 91.18 & 84.21 & 84.13 & 90.12 & 85.99 & 89.71 & 83.63 & 84.78 & 87.69 & 90.79 & 88.46 & 85.71 \\
claude-3-opus &82.65 & 78.67 & 85.57 & 91.18 & 85.96 & 78.84 & 81.07 & 85.35 & 83.82 & 78.36 & 83.33 & 83.08 & 90.79 & 82.31 & 82.86 \\
claude-3-sonnet &66.94 & 64.0 & 69.07 & 85.29 & 73.68 & 58.73 & 68.31 & 72.61 & 65.2 & 64.91 & 71.01 & 61.54 & 69.74 & 62.31 & 70.0 \\
cohere-aya-32b &74.91 & 70.67 & 79.38 & 82.35 & 77.19 & 74.6 & 70.37 & 82.17 & 72.06 & 72.51 & 77.54 & 80.0 & 85.53 & 69.23 & 71.43 \\
cohere-aya-8b &64.77 & 57.33 & 79.38 & 64.71 & 71.93 & 59.79 & 64.2 & 74.52 & 64.71 & 63.74 & 62.32 & 67.69 & 64.47 & 54.62 & 64.29 \\
deepseek-v3 &84.29 & 81.33 & 91.75 & 85.29 & 82.46 & 81.48 & 86.83 & 85.35 & 84.31 & 77.78 & 84.78 & 81.54 & 93.42 & 85.38 & 80.0 \\
deepseek-r1 &86.93 & 85.33 & 88.66 & 88.24 & 82.46 & 87.83 & 89.71 & 87.26 & 85.29 & 83.63 & 84.78 & 86.15 & 92.11 & 88.46 & 85.71 \\
gemma-2-27b &78.25 & 76.0 & 87.63 & 88.24 & 80.7 & 74.6 & 75.72 & 84.08 & 79.41 & 70.18 & 79.71 & 83.08 & 81.58 & 76.15 & 75.71 \\
gemma-2-9b &75.73 & 72.0 & 82.47 & 85.29 & 77.19 & 72.49 & 73.66 & 78.98 & 75.98 & 71.35 & 80.43 & 80.0 & 84.21 & 69.23 & 72.86 \\
gpt-3.5-turbo-0125 &70.11 & 68.0 & 83.51 & 73.53 & 73.68 & 67.2 & 67.49 & 75.8 & 68.14 & 60.82 & 67.39 & 78.46 & 77.63 & 70.77 & 70.0 \\
gpt-4o &88.8 & 90.67 & 89.69 & 97.06 & 85.96 & 85.71 & 89.71 & 88.54 & 87.25 & 86.55 & 89.13 & 90.77 & 93.42 & 90.77 & 88.57 \\
gpt-4o-mini &82.3 & 77.33 & 88.66 & 88.24 & 78.95 & 80.42 & 81.48 & 87.9 & 83.82 & 76.61 & 81.16 & 84.62 & 92.11 & 80.77 & 75.71 \\
o1 &89.62 & 90.67 & 89.69 & 85.29 & 85.96 & 90.48 & 90.53 & 88.54 & 89.71 & 87.72 & 89.86 & 93.85 & 92.11 & 92.31 & 82.86 \\
o1-mini &81.48 & 82.67 & 83.51 & 82.35 & 84.21 & 77.25 & 82.3 & 82.17 & 83.33 & 74.85 & 83.33 & 78.46 & 86.84 & 85.38 & 78.57 \\
o3-mini &83.41 & 81.33 & 82.47 & 85.29 & 85.96 & 83.07 & 84.77 & 86.62 & 82.35 & 77.19 & 83.33 & 86.15 & 90.79 & 85.38 & 77.14 \\
grok2 &86.87 & 91.89 & 87.67 & 90.91 & 96.3 & 82.0 & 85.1 & 89.62 & 88.08 & 82.11 & 92.77 & 84.62 & 90.54 & 84.51 & 92.31 \\
llama-3-70b &81.36 & 80.0 & 84.54 & 88.24 & 80.7 & 79.89 & 82.72 & 82.8 & 84.8 & 75.44 & 81.88 & 81.54 & 88.16 & 76.92 & 75.71 \\
llama-3.1-405b &85.17 & 81.33 & 89.69 & 91.18 & 85.96 & 78.84 & 88.89 & 86.62 & 87.75 & 81.29 & 83.33 & 90.77 & 86.84 & 82.31 & 84.29 \\
llama-3.1-70b &83.24 & 77.33 & 85.57 & 91.18 & 84.21 & 80.95 & 82.72 & 89.17 & 85.29 & 78.36 & 84.78 & 83.08 & 85.53 & 80.77 & 81.43 \\
llama-3.1-8b &70.22 & 61.33 & 83.51 & 76.47 & 77.19 & 66.67 & 67.08 & 82.8 & 71.08 & 61.4 & 74.64 & 73.85 & 73.68 & 60.0 & 67.14 \\
llama-3.3-70b &82.36 & 76.0 & 84.54 & 85.29 & 75.44 & 78.31 & 85.19 & 87.9 & 83.82 & 77.19 & 86.23 & 83.08 & 85.53 & 80.0 & 80.0 \\
mistral-7b &65.01 & 65.33 & 81.44 & 79.41 & 73.68 & 61.9 & 61.32 & 71.34 & 64.22 & 60.23 & 64.49 & 64.62 & 71.05 & 55.38 & 61.43 \\
mistral-large &82.83 & 78.67 & 84.54 & 91.18 & 84.21 & 78.31 & 85.6 & 86.62 & 79.41 & 82.46 & 81.16 & 83.08 & 90.79 & 80.0 & 84.29 \\
mistral-nemo &71.92 & 69.33 & 81.44 & 76.47 & 77.19 & 69.84 & 70.37 & 84.08 & 69.12 & 64.91 & 73.91 & 72.31 & 75.0 & 66.15 & 67.14 \\
mistral-small &78.96 & 76.0 & 87.63 & 85.29 & 84.21 & 77.78 & 77.37 & 85.35 & 78.43 & 70.18 & 77.54 & 83.08 & 88.16 & 73.08 & 80.0 \\
mixtral-8x22b &73.8 & 74.67 & 80.41 & 91.18 & 80.7 & 68.25 & 71.6 & 82.8 & 73.53 & 67.25 & 73.91 & 73.85 & 78.95 & 68.46 & 72.86 \\
mixtral-8x7b &73.92 & 72.0 & 84.54 & 82.35 & 77.19 & 69.31 & 75.31 & 81.53 & 69.61 & 70.18 & 68.84 & 70.77 & 81.58 & 75.38 & 68.57 \\
qwen-2.5-72b &83.18 & 78.67 & 85.57 & 91.18 & 82.46 & 84.66 & 85.6 & 86.62 & 80.39 & 74.85 & 84.78 & 86.15 & 89.47 & 82.31 & 78.57 \\
        \bottomrule        \end{tabular}}
    \caption{Accuracy (\%) for 29 tested models on \benchmark-Easy at sub-continent level. Human baseline is 92.4\% and the random baseline is 25\%.}
    \label{app: eval_result_models_mcq}
    \label{fig: eval_result_models_mcq}

\end{table*}

\clearpage
\begin{table*}[ht]
\resizebox{1.1\textwidth}{!}{
   \centering
    \begin{tabular}{c|c|c|c|c|c|c|c|c|c|c|c|c|c|c|c}
            \toprule
\textbf{\makecell[c]{Model}} & \textbf{\makecell[c]{Overall}}  & \textbf{\makecell[c]{North\\Africa}} &  \textbf{\makecell[c]{South\\Africa}} &  \textbf{\makecell[c]{West\\Africa}} &  \textbf{\makecell[c]{North\\America}} &  \textbf{\makecell[c]{South\\America}} &  \textbf{\makecell[c]{East\\Asia}} &  \textbf{\makecell[c]{South\\Asia}} &  \textbf{\makecell[c]{Southeast\\Asia}} &  \textbf{\makecell[c]{Middle East/\\West Asia}} &  \textbf{\makecell[c]{East\\Europe}} &  \textbf{\makecell[c]{North\\Europe}} &  \textbf{\makecell[c]{South\\Europe}} &  \textbf{\makecell[c]{West\\Europe}} &  \textbf{\makecell[c]{Oceania}} \\
\midrule
claude-3.5-haiku &50.12 & 49.33 & 55.67 & 70.59 & 64.91 & 42.33 & 54.73 & 56.05 & 48.04 & 44.44 & 37.68 & 55.38 & 57.89 & 47.69 & 48.57 \\
claude-3.5-sonnet &53.46 & 46.67 & 57.73 & 70.59 & 64.91 & 49.21 & 58.44 & 57.96 & 52.45 & 42.69 & 42.75 & 58.46 & 65.79 & 53.85 & 52.86 \\
claude-3-opus &52.87 & 54.67 & 61.86 & 55.88 & 59.65 & 41.8 & 56.79 & 62.42 & 55.39 & 44.44 & 50.0 & 66.15 & 51.32 & 44.62 & 50.0 \\
claude-3-sonnet &47.25 & 44.0 & 54.64 & 52.94 & 56.14 & 42.33 & 43.21 & 52.23 & 47.55 & 47.95 & 45.26 & 42.42 & 51.32 & 45.38 & 51.43 \\
cohere-aya-32b &31.18 & 25.33 & 35.05 & 38.24 & 54.39 & 34.39 & 28.4 & 35.67 & 30.39 & 18.71 & 36.96 & 29.23 & 26.32 & 30.77 & 30.0 \\
cohere-aya-8b &28.66 & 25.33 & 29.9 & 35.29 & 40.35 & 32.28 & 29.63 & 33.12 & 30.73 & 25.73 & 26.28 & 26.15 & 13.16 & 23.85 & 28.57 \\
deepseek-v3 &38.86 & 32.0 & 49.48 & 41.18 & 50.88 & 40.21 & 44.44 & 37.58 & 38.73 & 23.39 & 34.06 & 47.69 & 47.37 & 34.62 & 38.57 \\
deepseek-r1 &57.33 & 46.67 & 56.7 & 61.76 & 57.89 & 56.08 & 61.32 & 59.87 & 53.43 & 51.46 & 61.59 & 60.0 & 71.05 & 54.62 & 55.71 \\
gemma2-27b &42.56 & 29.33 & 54.64 & 61.76 & 61.4 & 40.21 & 46.09 & 51.59 & 35.29 & 35.67 & 34.06 & 60.0 & 39.47 & 40.77 & 34.29 \\
gemma2-9b &41.44 & 25.33 & 52.58 & 64.71 & 59.65 & 37.57 & 44.44 & 53.5 & 38.73 & 33.33 & 34.06 & 50.77 & 39.47 & 35.38 & 37.14 \\
gpt-3.5-turbo-0125 &34.53 & 29.33 & 34.02 & 47.06 & 54.39 & 34.92 & 32.1 & 42.04 & 31.86 & 23.39 & 28.26 & 43.08 & 39.47 & 39.23 & 34.29 \\
gpt-4o &60.49 & 54.67 & 68.04 & 73.53 & 71.93 & 57.14 & 64.61 & 68.79 & 59.31 & 49.12 & 52.17 & 66.15 & 69.74 & 56.15 & 57.14 \\
gpt-4o-mini &46.78 & 41.33 & 53.61 & 50.0 & 61.4 & 42.33 & 51.03 & 55.41 & 46.08 & 34.5 & 38.41 & 55.38 & 48.68 & 47.69 & 44.29 \\
o1 &61.37 & 54.67 & 62.89 & 52.94 & 57.89 & 51.32 & 70.78 & 61.78 & 63.24 & 58.48 & 61.59 & 66.15 & 61.84 & 61.54 & 62.86 \\
o1-mini &49.59 & 41.33 & 54.64 & 50.0 & 59.65 & 47.62 & 53.09 & 58.6 & 48.04 & 40.35 & 43.48 & 53.85 & 53.95 & 46.92 & 51.43 \\
o3-mini &56.51 & 52.0 & 53.61 & 70.59 & 68.42 & 50.79 & 63.37 & 61.78 & 59.31 & 43.27 & 50.0 & 61.54 & 59.21 & 57.69 & 55.71 \\
grok2 &54.0 & 34.38 & 47.46 & 66.67 & 44.44 & 50.0 & 59.57 & 61.11 & 59.09 & 49.49 & 56.94 & 56.82 & 50.88 & 50.77 & 54.17 \\
llama-3-70b &47.13 & 42.67 & 52.58 & 61.76 & 63.16 & 45.5 & 50.62 & 55.41 & 43.63 & 40.35 & 40.58 & 60.0 & 39.47 & 40.77 & 45.71 \\
llama-3.1-405b &51.93 & 37.33 & 63.92 & 73.53 & 64.91 & 48.15 & 56.38 & 59.87 & 50.0 & 40.94 & 47.1 & 61.54 & 55.26 & 43.85 & 51.43 \\
llama-3.1-70b &54.57 & 46.67 & 63.92 & 50.0 & 64.91 & 49.21 & 62.96 & 58.6 & 51.47 & 47.95 & 48.55 & 60.0 & 63.16 & 48.46 & 54.29 \\
llama31-8b &35.99 & 26.67 & 48.45 & 41.18 & 59.65 & 35.45 & 37.45 & 42.04 & 34.31 & 26.32 & 31.16 & 43.08 & 42.11 & 26.15 & 32.86 \\
llama-3.3-70b &54.4 & 48.0 & 69.07 & 58.82 & 66.67 & 50.26 & 62.55 & 57.32 & 47.06 & 48.54 & 46.38 & 61.54 & 61.84 & 48.46 & 52.86 \\
mistral-7b &33.94 & 20.0 & 44.33 & 47.06 & 49.12 & 33.86 & 34.16 & 43.31 & 31.86 & 25.15 & 31.16 & 43.08 & 27.63 & 30.0 & 32.86 \\
mistral-large &47.6 & 38.67 & 49.48 & 35.29 & 59.65 & 41.27 & 52.67 & 45.86 & 53.92 & 40.94 & 49.28 & 53.85 & 52.63 & 44.62 & 42.86 \\
mistral-nemo &36.05 & 29.33 & 47.42 & 35.29 & 61.4 & 26.46 & 35.8 & 42.68 & 34.8 & 30.41 & 31.88 & 47.69 & 43.42 & 29.23 & 38.57 \\
mistral-small-3 &41.97 & 38.67 & 46.39 & 32.35 & 59.65 & 26.46 & 46.91 & 45.86 & 40.2 & 35.67 & 45.65 & 53.85 & 53.95 & 39.23 & 40.0 \\
mixtral-8x22b &44.72 & 40.0 & 45.36 & 52.94 & 59.65 & 42.33 & 40.57 & 51.59 & 46.57 & 39.18 & 40.15 & 44.62 & 42.11 & 50.0 & 48.57 \\
mixtral-8x7b &21.16 & 13.33 & 24.74 & 35.29 & 26.32 & 15.34 & 25.93 & 27.39 & 18.63 & 16.37 & 15.94 & 16.92 & 26.32 & 23.85 & 21.43 \\
qwen25-72b &46.72 & 38.67 & 49.48 & 44.12 & 56.14 & 38.1 & 58.85 & 53.5 & 46.57 & 39.18 & 41.3 & 55.38 & 48.68 & 42.31 & 38.57 \\
        \bottomrule
        \end{tabular}}
    \caption{Accuracy (\%) for 29 tested models on \benchmark-Hard at sub-continent level. Human baseline is 92.4\% and the random baseline is $(\frac{1}{2})^{4} = 6.25\%$.}
    \label{app: eval_result_models_binary}
    \label{fig: eval_result_models_binary}
\end{table*}

\clearpage
\section{\method: AI-assisted Red teaming System}
\label{app: red-teaming-demo-detail}

\begin{figure*}[h!]
    \centering
\includegraphics[width=\textwidth]{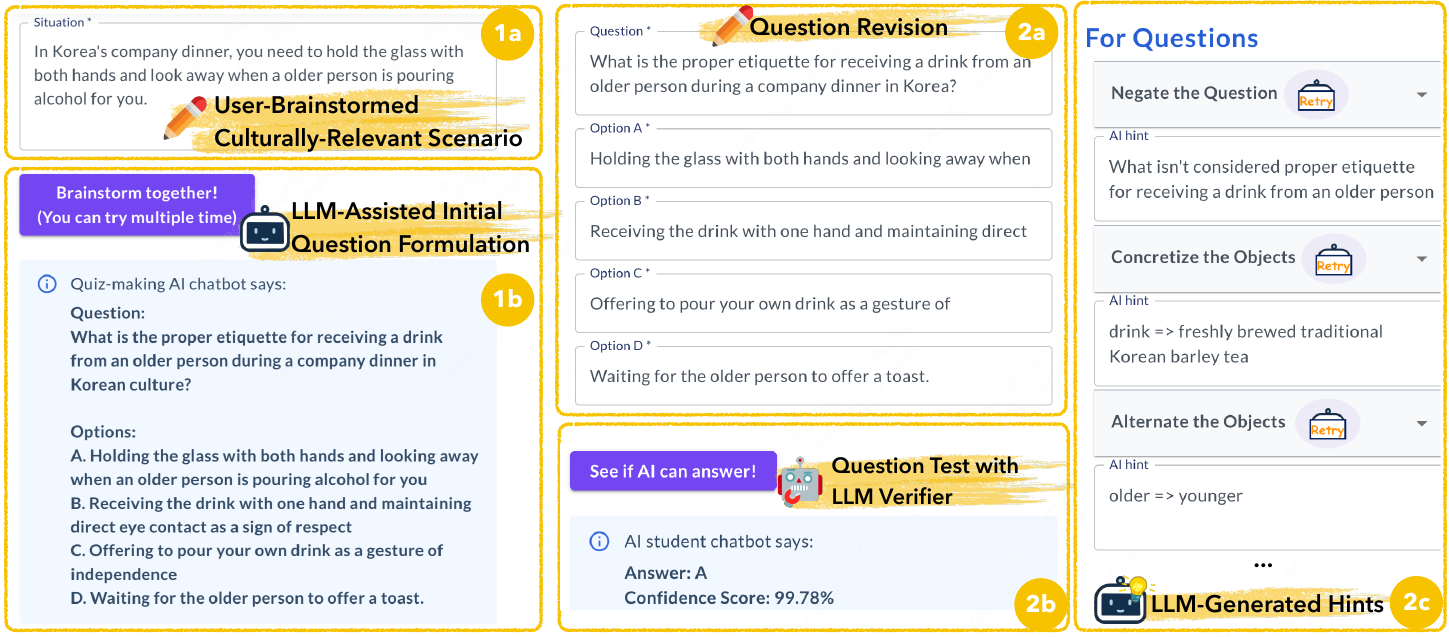}
\vspace{-0.4cm}
    \caption{Interface for Step 1 (Data collection via human-AI collaborative Red teaming). \textbf{(1a)} Users brainstorm culturally relevant scenarios \textbf{(1b)} Users convert scenarios to MCQs with LLM-powered Question Formulation \textbf{(2a)} Users revise MCQs and \textbf{(2b)} Test MCQs based on the chosen option and its confidence score from LLM Verifier \textbf{(2c)} Users inspire by LLM-generated hints with strategies e.g., Negation, Synonym.}
    \label{fig:red_teaming_demo}
\end{figure*}

\textbf{Step 1: Data collection via human-AI collaborative red teaming}

This system consists of two steps, as demonstrated in Fig. \ref{fig:red_teaming_demo} -- 1) Question Formulation 2) Question Verification and Revision 3) Feedback Collection. The first two steps involve a red-teaming exercise to formulate a challenging question step-by-step.

\begin{figure*}[h]
\centering
\includegraphics[width=0.9\textwidth]{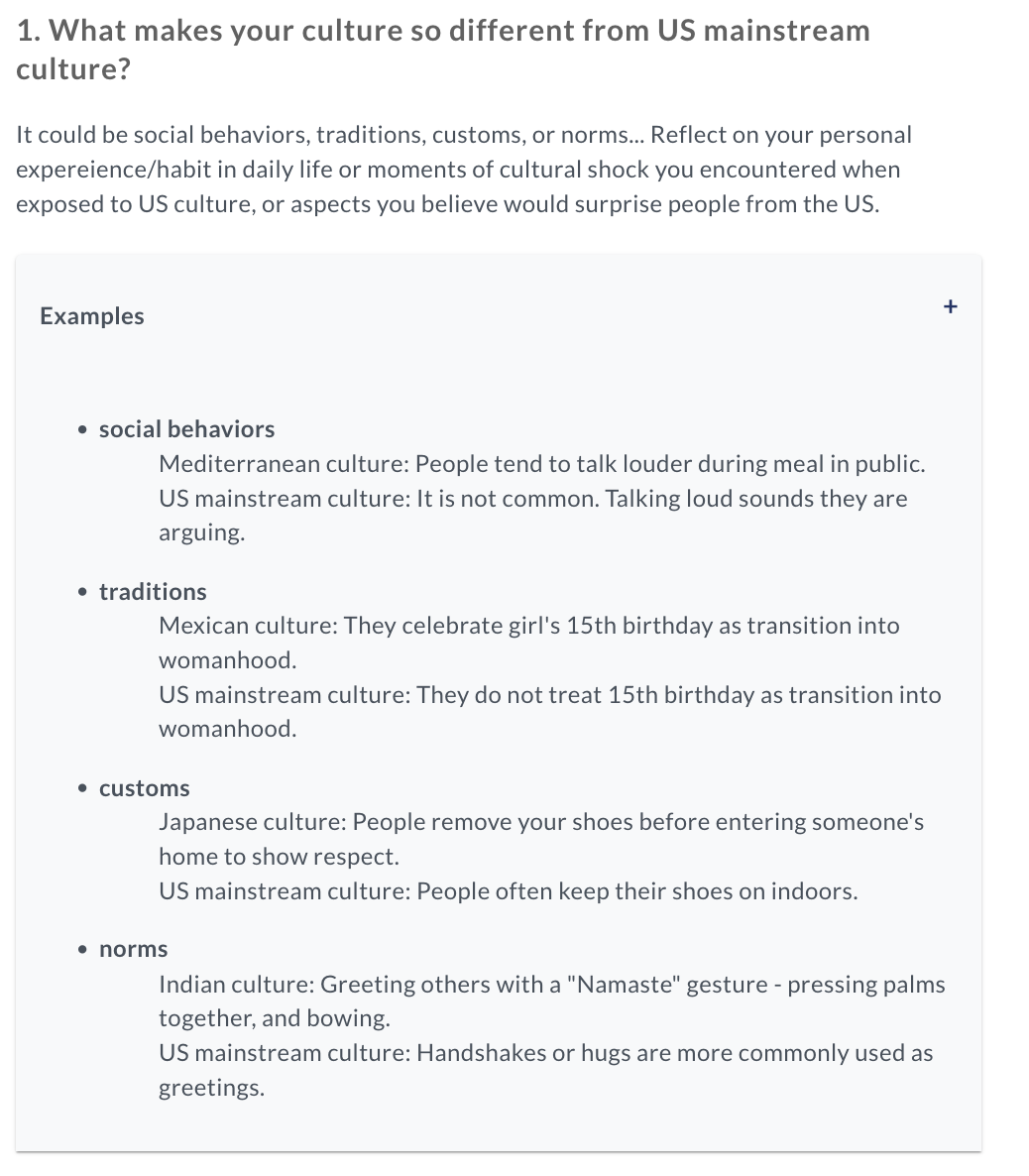}
\caption{Guidance on Step 1a (brainstorming culturally-relevant scenario) in our human-AI collaborative red teaming system (Part 1).}
\label{fig:guidance-step-1}
\end{figure*}

\begin{figure*}[h]
\centering
\includegraphics[width=0.9\textwidth]{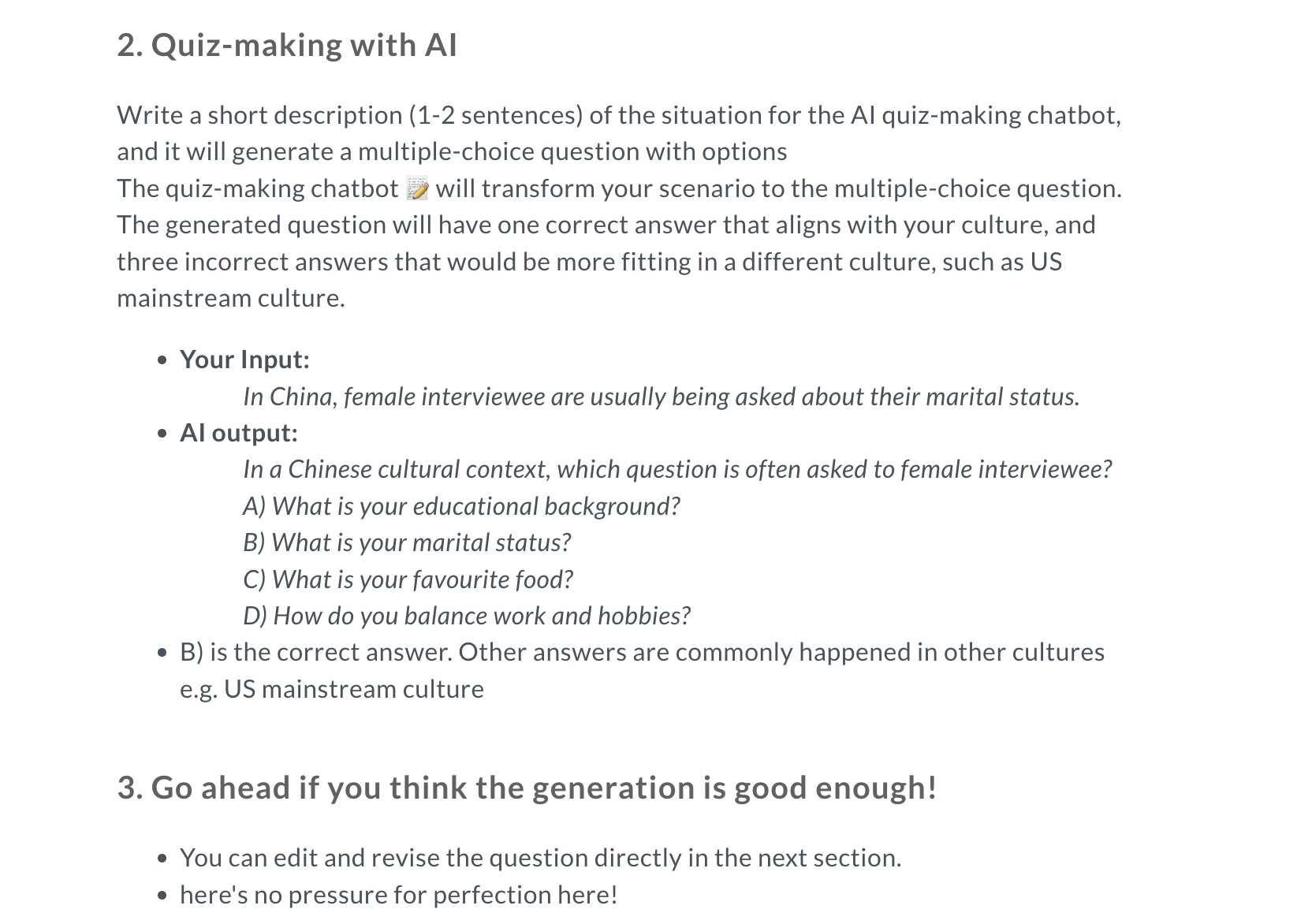}
\caption{Guidance on Step 1a (brainstorming culturally-relevant scenario) in our human-AI collaborative red teaming system (Part 2).}
\label{fig:guidance-step-1-2}
\end{figure*}

\begin{figure*}[h]
\centering
\includegraphics[width=0.9\textwidth]{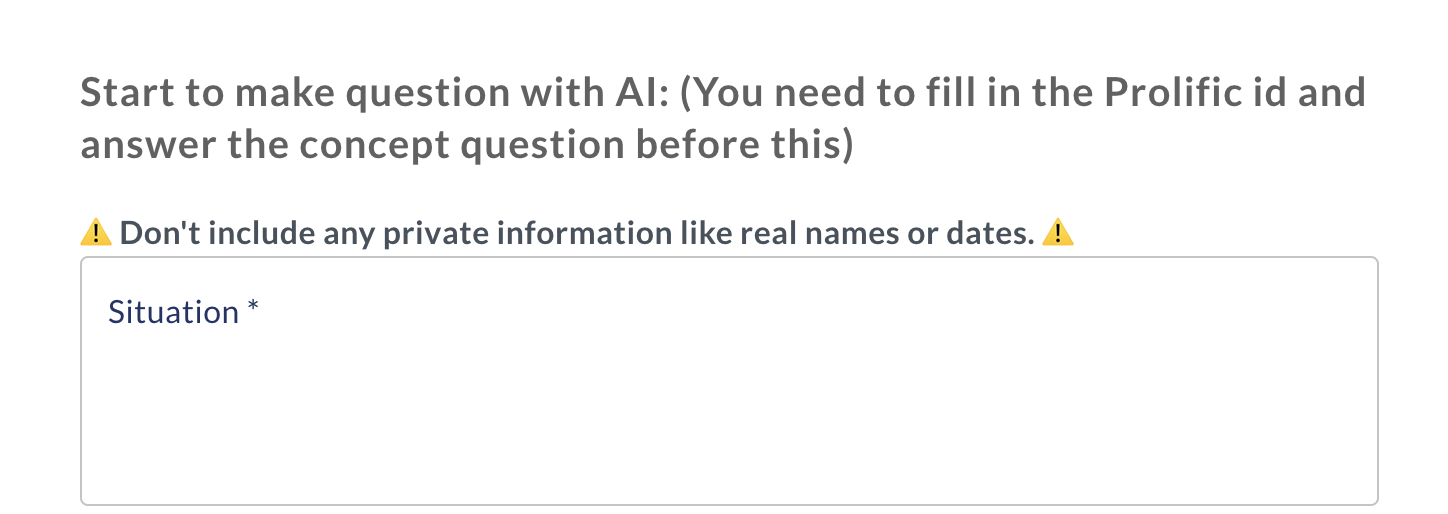}
\caption{Guidance on Step 1a (brainstorming culturally-relevant scenario) in our human-AI collaborative red teaming system (Part 3).}
\label{fig:guidance-step-1-3}
\end{figure*}

\paragraph{Step 1a: Question Formulation.} The goal is to facilitate users in brainstorming culturally relevant situations based on their personal experiences. A step-by-step guideline with detailed examples is provided to inspire them, as shown in Fig. \ref{fig:guidance-step-1}, \ref{fig:guidance-step-1-2} and \ref{fig:guidance-step-1-3}. Users formulate a multiple-choice question (MCQ), which comprises one correct and culturally appropriate option.

\begin{figure}[ht]
\centering
\includegraphics[width=0.5\textwidth]{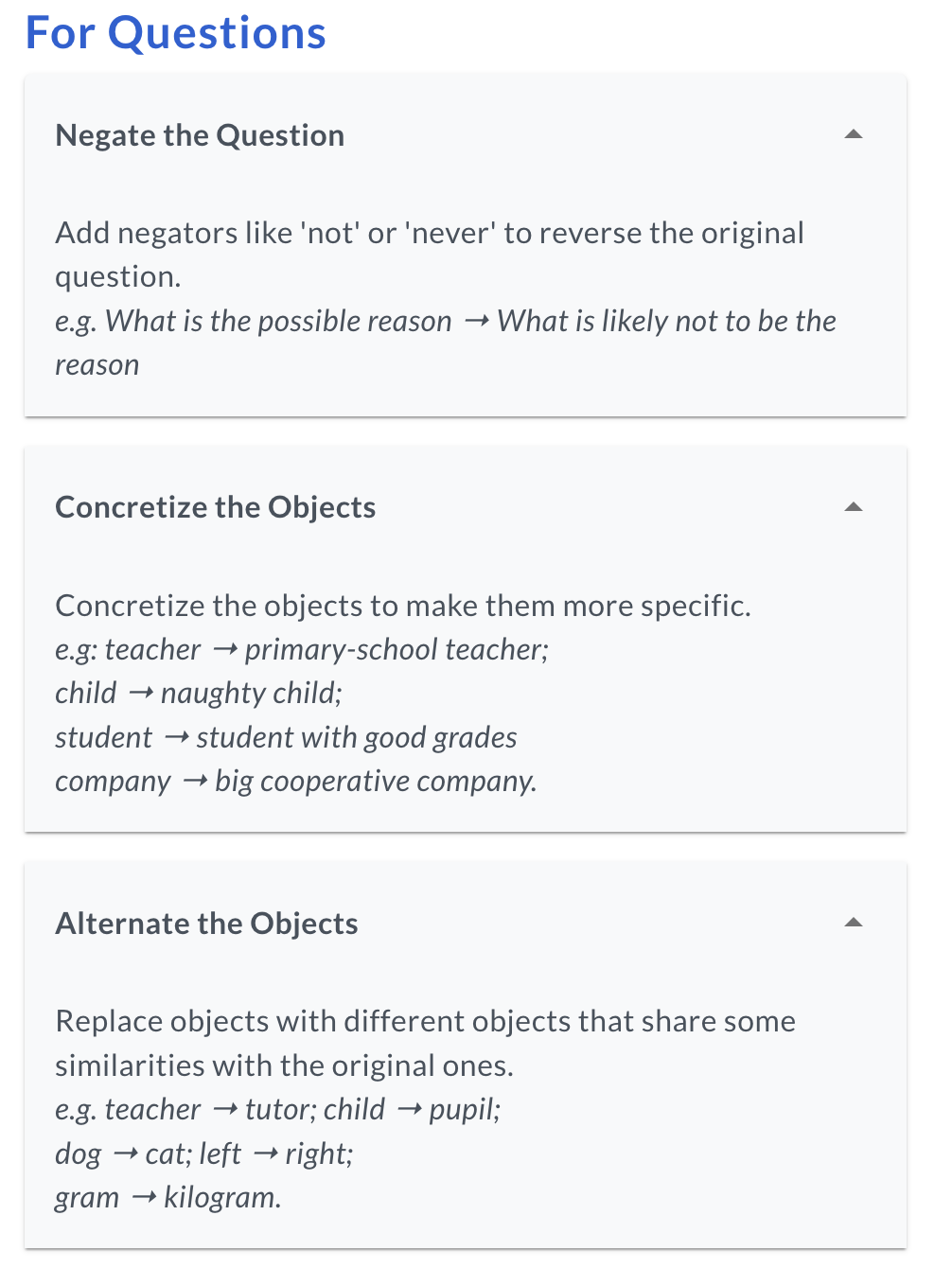}
\includegraphics[width=0.5\textwidth]{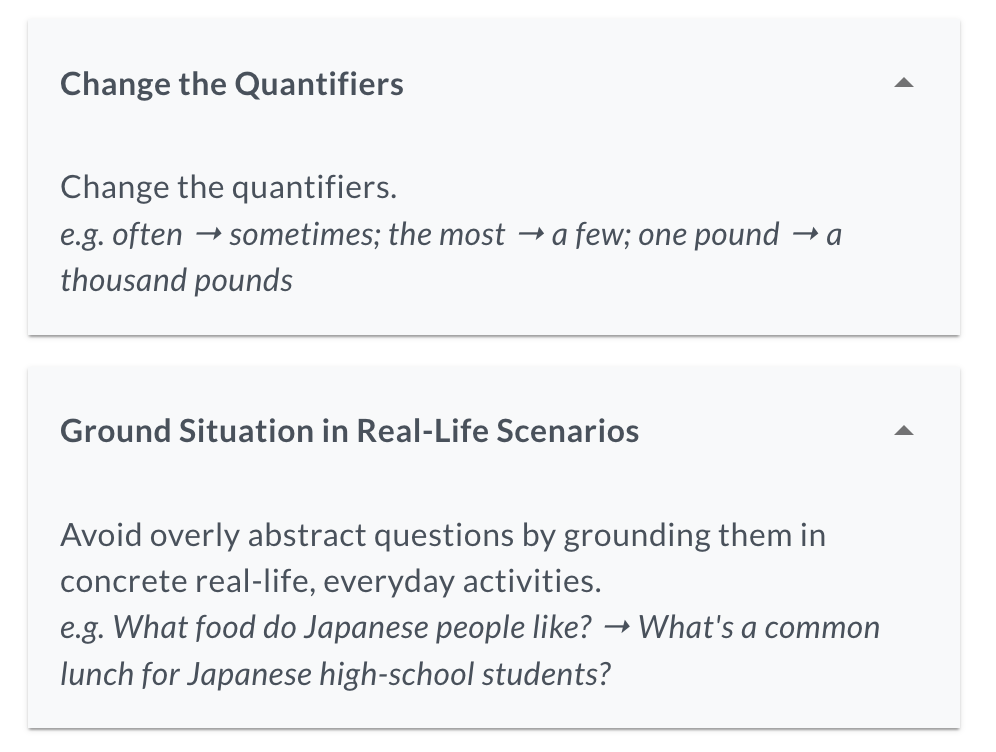}
\includegraphics[width=0.5\textwidth]{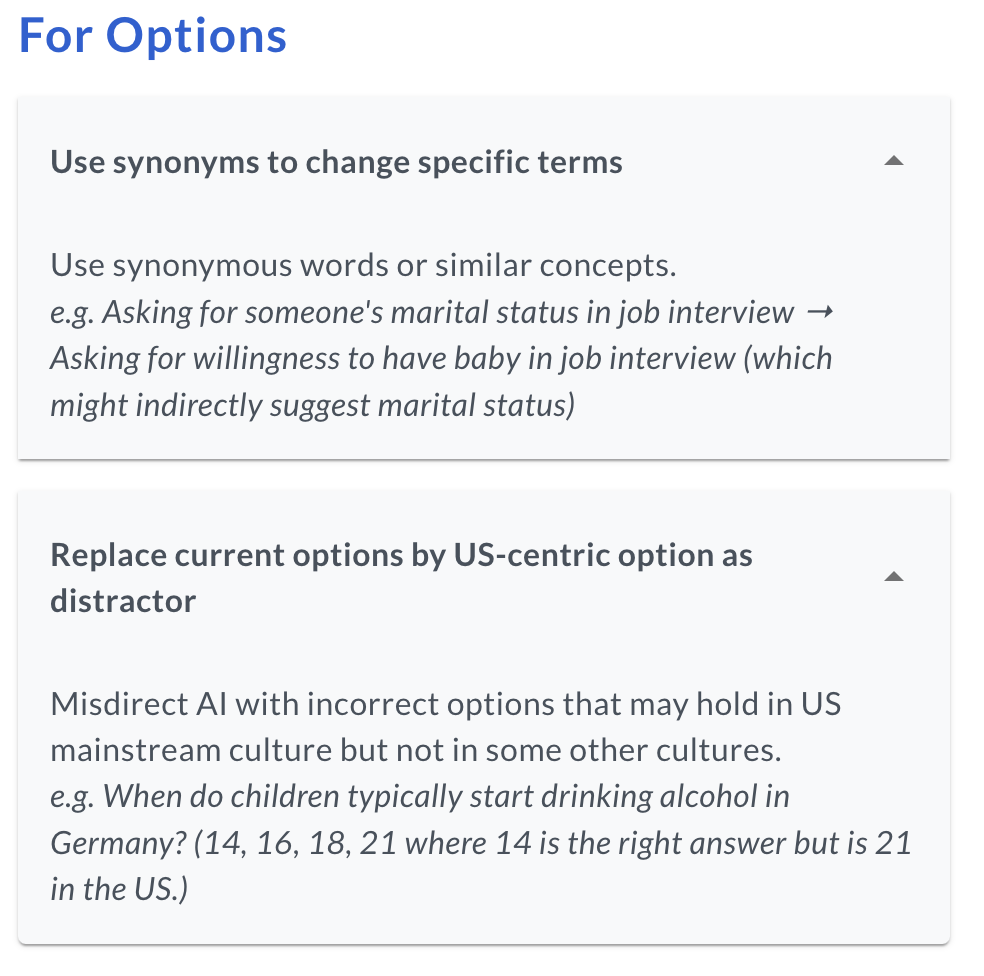}
\caption{Guidance on Step 1b (Question verification \& revision) in our human-AI collaborative red teaming system.}
\label{fig:guidance-step-2}
\end{figure}

\paragraph{Step 1b: Question Verification \& Revision.}
This step provides an interactive and iterative red-teaming platform that allows users to verify their culturally sensitive MCQs. The platform assists them in revising the question and the options to make it more challenging by providing descriptions of various common revision strategies with drafted examples (e.g., ``Negate the Question''), as stated in Fig. \ref{fig:red_teaming_demo} and Fig. \ref{fig:guidance-step-2}.

\clearpage

\clearpage

\clearpage

\clearpage

\clearpage
\clearpage
\begin{figure*}[h]
\centering
\includegraphics[width=0.9\textwidth]{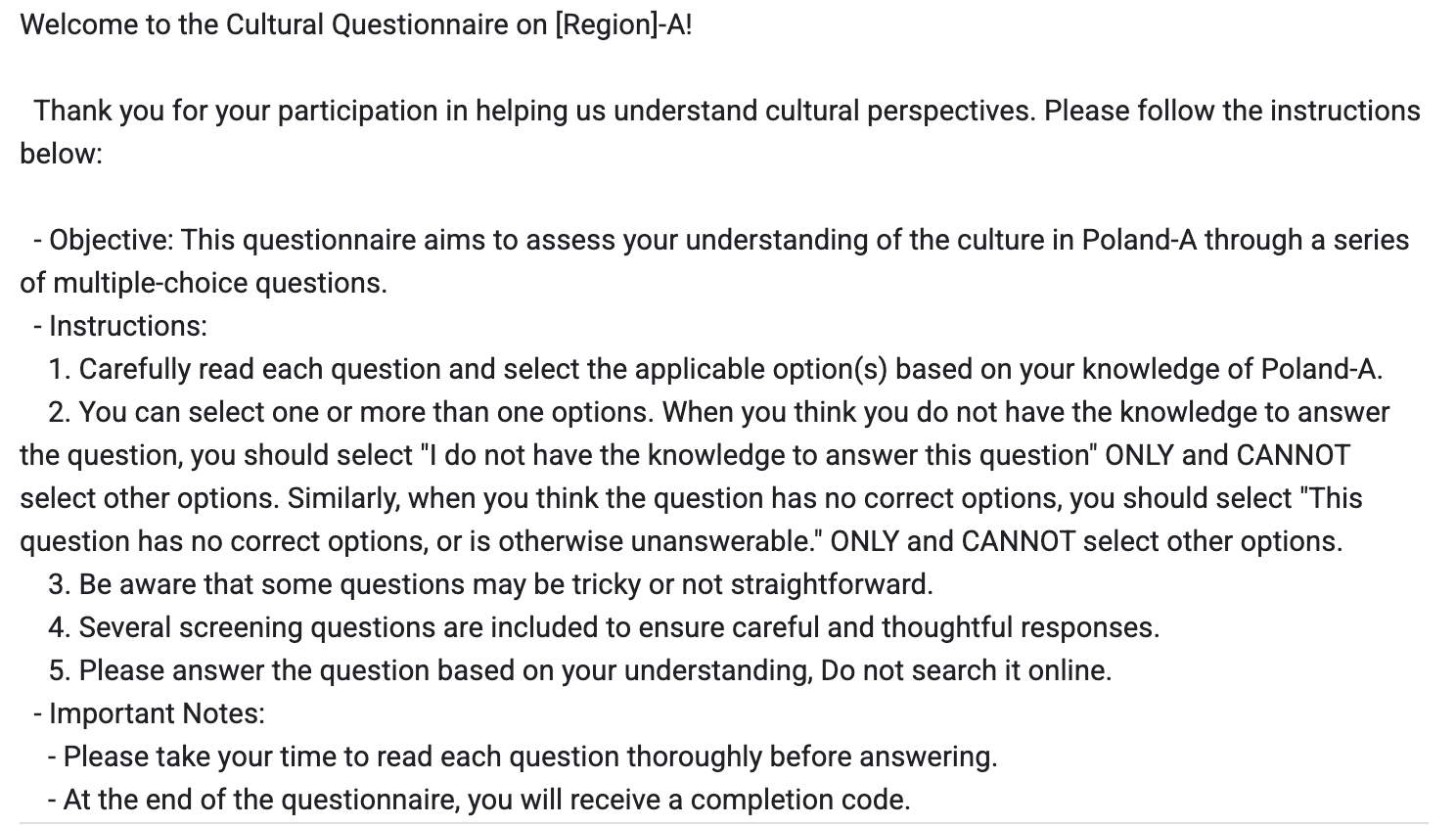}
\caption{Detailed instructions for Step 2 (quality check) after collecting our data on the human-AI red teaming platform.}
\label{fig:qc-guidance-step-1}
\end{figure*}
\clearpage
\begin{figure*}[h]
\centering
\includegraphics[width=0.9\textwidth]{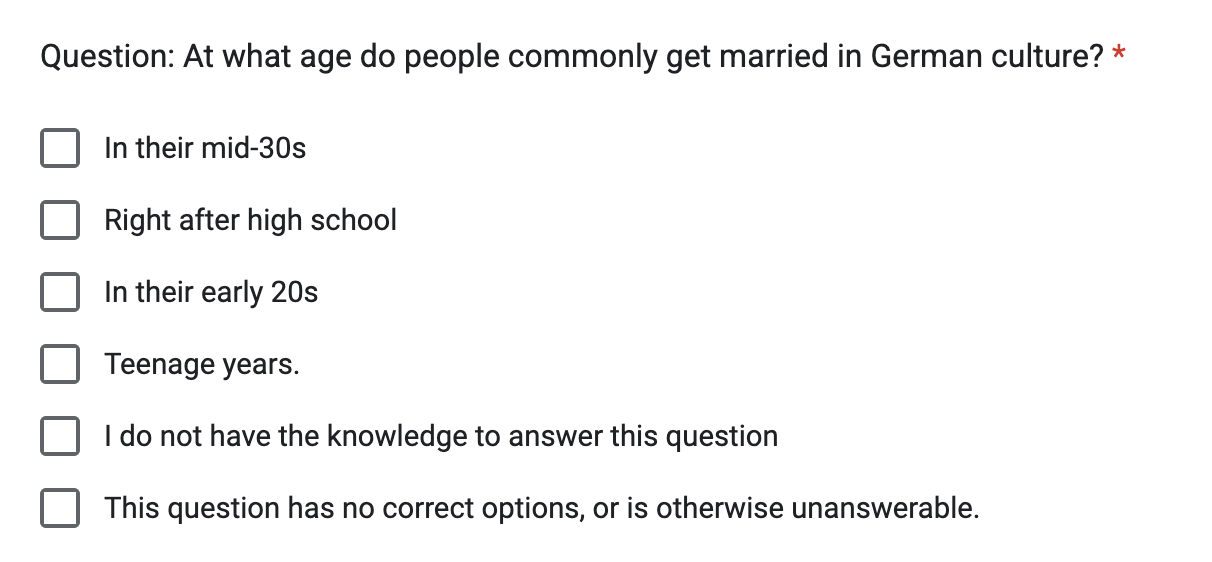}
\caption{Sample Question for Step 2 (quality check). Participants can select multiple options to indicate all the possible answers per question. They can also select ``I do not have the knowledge to answer this question" and ``This question has no correct options, or is otherwise unanswerable." that allow us to filter out unsuitable questions.}
\label{fig:qc-sample-question}
\end{figure*}

\clearpage
\begin{figure*}[h]
    \centering
\resizebox{\textwidth}{!}{%
\includegraphics[width=\textwidth]{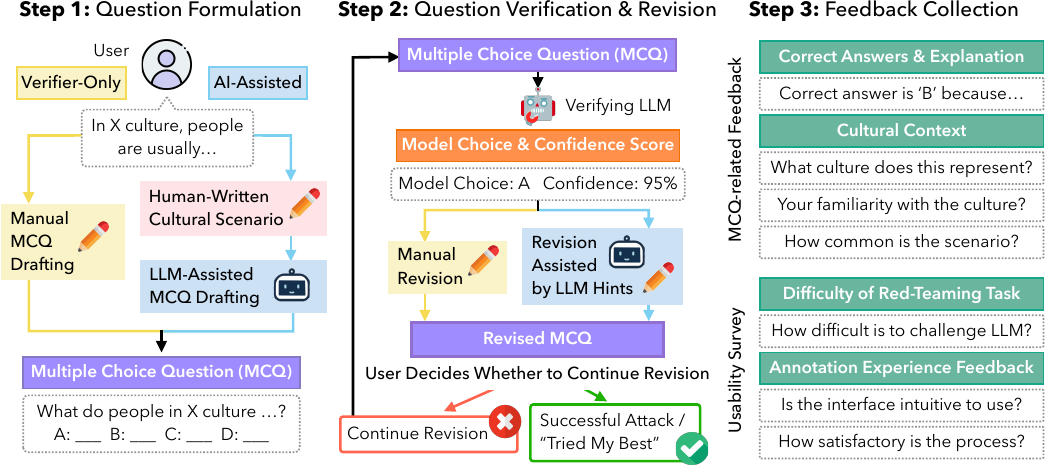}}
    \vspace{-0.4cm}
    \caption{Two settings of \methodemoji\method (1) \hlc[workflow-variant-1-color]{Verifier-Only} (2)  \hlc[workflow-variant-2-color]{AI-Assisted}. \textbf{Step 1:} Users brainstorm a culturally relevant scenario and use it to draft a multiple-choice question (MCQ). In (1), users manually draft the MCQ. In (2), an LLM drafts an MCQ based on a user-provided seed scenario. \textbf{Step 2:} Users test the question with the model and revise it iteratively until satisfied. In (1), users manually revise the MCQ. In (2), users revise with hints from an LLM. \textbf{Step 3:} Users provide gold answers and feedback.}
    \label{fig:workflow}
\end{figure*}

\section{Pivot study on \method: Two settings in Human-AI Collaborations for red-teaming in cultural knowledge}

\subsection{Varying Levels of AI Assistance: \variantone \& \varianttwo}

To study the effect of different AI-assistance modules, we have implemented two settings with varying AI Assistance levels in our experiment: (1) \variantone (2) \varianttwo, as presented in Fig. \ref{fig:workflow}. The first setting includes the basic element (Verifier) for the red-teaming exercise. The second setting aims to explore question formulation and hints on revision powered by LLMs. 

\textbf{\variantone} The users formulate their MCQ in Step 1 (Question formulation). They then present their MCQ to LLM Verifier and revise their question iteratively by themselves. More specifically, the Verifier responds with one option coupled with the corresponding confidence score (which is the top-1 linear probability of its log probability). This aids users in judging the question's difficulty. In our system, GPT-3.5-turbo (0125) served as the Verifier to reduce cost and time latency  \citep{brown2020language}.

\textbf{\varianttwo} Users benefit from extensive AI-assistance in question formation and revision, as illustrated in Fig. \ref{fig:workflow}. In Step 1, users convert their short cultural-relevant scenarios into MCQs using LLM-powered question formation. In Step 2, users verify and revise their questions with LLM-powered suggestions. The Verifier remains consistent with the \variantone setting.

\subsection{User Experiments with Red-teaming Workshops}

We recruit annotators from an academic institution to participate in our 1-hour workshop. Annotators are randomly assigned to use either \variantone or \varianttwo. After finishing the self-contained tutorial, we encourage annotators to interact with the system for at least 45 minutes, allowing them to leave at their own will. 

\clearpage
\subsection{Exploring a better Human-AI collaboration on \benchmark-v0.1}
\begin{figure*}[h!]

    \begin{minipage}{0.5\textwidth}
        \centering
        \includegraphics[width=\textwidth]{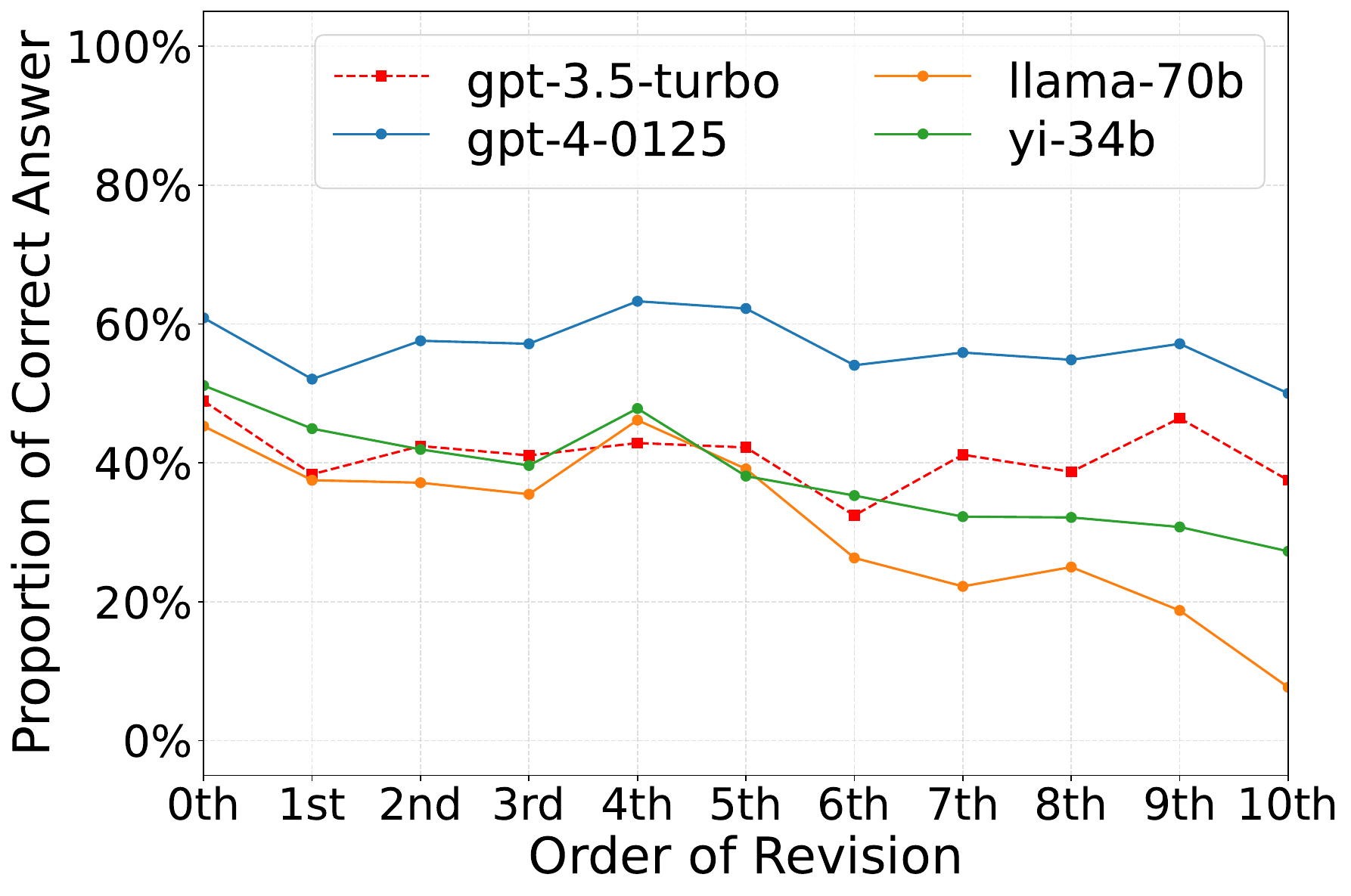}
    \end{minipage}
    \begin{minipage}{0.5\textwidth}
        \centering
\includegraphics[width=\textwidth]{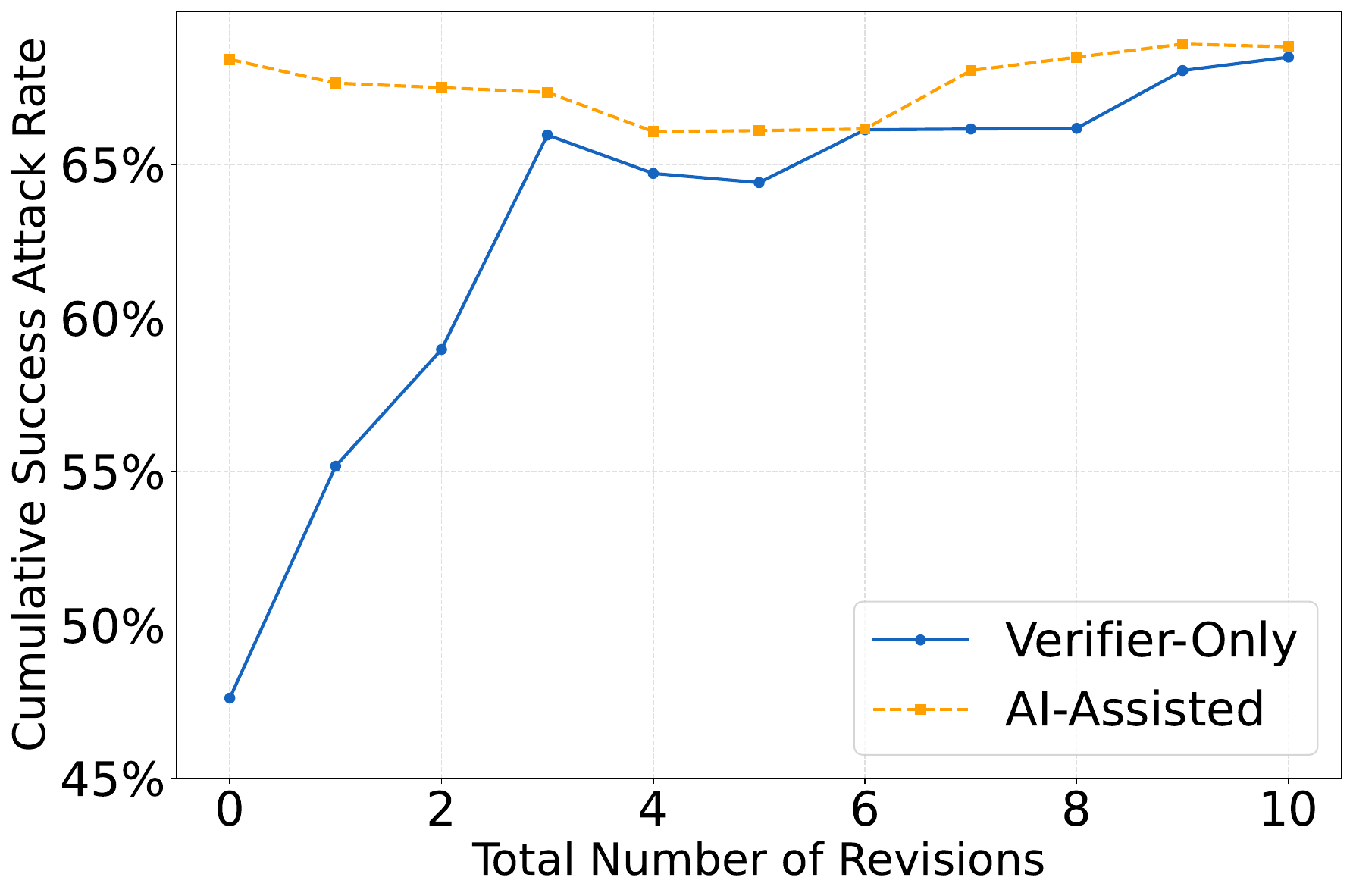}

    \end{minipage}
\caption{LLM assistance on (1) \textbf{Left}: Verifier (\textcolor[HTML]{fc4c4c}{gpt-3.5-turbo}) by comparing other models performance on questions by users without other LLM assistance (\variantone) (2) \textbf{Right}: Revision by comparing between the final success attack rate and the total number of edits between users with LLM Hints (\textcolor[HTML]{f4800b}{\varianttwo}) and without LLM Hints (\textcolor[HTML]{404ba4}{\variantone}).}

\label{fig:combined-ablation-graphs}
\end{figure*}

We collected 252 carefully-reviewed MCQs spanned across 34 diverse cultures. After data collection, we manually reviewed the questions to ensure they 1) follow the MCQ format 2) specify what kind of culture the question is asking for. 

With this \benchmark-v0.1, we explore and provide insights on how to make use of human-AI collaboration to produce a challenging dataset on multicultural knowledge. We analyze (1) AI-assistance contribution to different system modules, (2) behavior and perception of users with varying levels of AI-assistance.

\subsection{Analysis 1: How does AI-assistance contribute to each module in our system?}

\subsubsection{LLM assistance on question formulation}
Question formulation entails brainstorming a culturally relevant scenario and transforming it into a multiple choice question (MCQ). Users with LLM-powered question formulation are only tasked to provide a sentence-length scenario description, whereas users without LLM-assistance need to formulate the MCQ question and answer options themselves. We first compare the time difference in the question formulation between the two groups.

\textbf{LLM assistance on question formulation \textit{does not significantly improve} the time needed on formulating questions.} The time needed for the initial MC questions created by \variantone and \varianttwo settings show no significant difference. Specifically, the former needs 183.4 seconds (SD=105.2) on average while the latter needs 161.8 (SD=109.4) on average.

Our formulation of red teaming tasks by writing structured MCQs appears to be relatively easy for human users without AI assistance. They effectively create their initial question template and attain equally impressive performance in terms of efficiency. One possible reason could be users without AI assistance formulate both questions and options concurrently when they are brainstorming. As we move forward in our future work, we should also consider how AI can aid people in brainstorming culturally relevant scenarios, rather than solely helping them in formulating structured questions.

\subsubsection{LLM assistance on Question Verifier}
After the MCQ is initially formulated, we attempt to use GPT-3.5-turbo as our Verifier \citep{brown2020language}. Users revise and verify the MCQ iteratively based on the Verifier response.
In this section, we evaluate the viability of this setup by investigating whether the Verifier serves as a good estimator for the degree of challenge for other LLMs. We assume that users revise their MCQs based on the Verifier response and its corresponding confidence score.

\textbf{Our LLM Verifier \textit{is able to create challenging questions} for all tested models.} To delineate the effect of LLM assistance, we analyze the MCQs created by users without LLM assistance on other LLMs in Fig. \ref{fig:combined-ablation-graphs} (Left). We found that LLM performance decreases with the number of revisions, especially for llama-70b by 38.73\% and yi-34b by 23.82\%. GPT-3.5-turbo (our Verifier model) decreased from 48.91\% to 37.5\%. GPT-4-turbo, which is the best performance model among all, decreased from 60.87\% to 50.00\% on the 10th revision. This shows that our LLM Verifier is capable of increasing the difficulty of questions for most models, as the number of revisions increases based on the response of LLM Verifier.

It implies that using only one model (GPT-3.5-turbo) as the Verifier can effectively construct challenging data for most of the models. However, if the goal of future work is to improve the stronger model (GPT-4-turbo), the relatively small decrease in it suggests that we need to randomly select different models as verifier, including GPT-4 model to help constructing a more challenging dataset.

\subsubsection{LLM assistance on Question Revision}
GPT-4 is used to generate LLM-powered hints based on the work-in-progress MCQ to provide revision suggestions with common revision strategies. Users without LLM-powered hints, on the other hand, are provided with static descriptions of those strategies. We now investigate the relationship between the number of revisions and the success attack rate of the finalized question (final success rates) for both users with LLM-assistance (\varianttwo setting) and users without LLM-assistance (\variantone settings).

\textbf{LLM assistance on Hints helps users to construct \textit{more challenging questions} after several rounds of revision.} 
For the initial three revisions, the final success attack rate of questions without LLM-powered suggestions (\variantone) increases. Conversely, questions with LLM-powered suggestions (\varianttwo) remain steady as they are initially created by the LLM-powered question formulation and are of good quality. Beyond four revisions, questions created in \varianttwo begin to show an increasing trend. On the other hand, those in \variantone require more revisions to match similar performance at 10 revisions, relative to those in \varianttwo. We can see the potential benefits of LLM suggestions become more apparent, especially after several rounds of revisions.

One reason is that individuals start with their initial ideas, which may deplete after a few attempts. LLM suggestions can therefore serve as an idea pool, providing diverse editing ideas that are grounded in their input question template. This illustrates the use of AI in facilitating creative thought and efficient question revision among individuals.

\subsection{Analysis 2: How users' behaviors and perception differ with varying levels of AI-assistance}

We compare and analyze annotators' behaviors and their perception. We used a two-sided Student t-test for the following analysis.
  \begin{table*}[h!]
  
    \centering
    \begin{tabular}{|c|c|c|c|} 
     \hline
     \textbf{Type} &\textbf{Verifier-Only} & \textbf{AI-Assisted} & $\Delta$\\
     \hline
     \multicolumn{4}{|c|}{\textit{\textbf{Whole Question Template Creation}}}\\
    \hline
   All questions created &$426.8_{163.6}$& $605.1_{404.0}$ & $178.2$\\
     \hline
    Questions that successfully attack & $426.4_{181.1}$& $590.9_{413.3}$ & $164.45$\\  
     \hline
   Questions trials after the first time &$461.25_{279.4}$ & $520.6_{348.4}$ & $59.35$\\  
    \hline
    \multicolumn{4}{|c|}{\textit{\textbf{Initial Question Template Formulation}}}\\
     \hline
    All questions created & $183.4_{105.2}$& $161.8_{109.4}$& -21.6\\

    \hline
   Questions that successfully attack &$1181.9_{107.9}$& $164.5_{107.4}$&$-17.4$\\
     \hline
    Questions trials after the first time& $195.6_{135.6}$& $170.3_{147.3}$ & $-25.3$\\
\hline
     \multicolumn{4}{|c|}{\textit{\textbf{Revision}}}\\
\hline
   All questions created & $95.8_{94.0}$& $248.2_{260.0}$ & $152.4^{*}$\\
    \hline
   Questions that successfully attack &  $101.3_{101.2}$& $235.8_{263.2}$ & $134.55$\\ 
     \hline
    Questions trials after the first time &$161.9_{188.7}$& $236.4_{196.9}$ & $74.48$\\
\hline
    \end{tabular}
    \caption{Average time taken per question (in seconds) for question creation both variants at $p=0.1$.}
    \label{table:results-time-taken-combined}
    \end{table*}
\subsubsection{Behaviors}
\label{sec:user-behavior}

\textbf{Users with more AI-assistance spend \textit{more time} on revising and make \textit{more revisions}, compared with those with less AI-assistance.} 
As shown in Table \ref{table:results-time-taken-combined}, users with LLM-generated hints (\varianttwo setting) require 152.4 more seconds, compared to users without LLM-generated Hints  (\variantone) with $p=0.1$. 
Similarly, users in \varianttwo setting also make 7.19 more revisions than users in \variantone. One possible reason is that the LLM-generated hints provide more information and inspiration for users to revise their questions. Users with LLM-generated hints tend to try and adopt various existing hints whereas users without LLM-generated hints tend to give up on revising and restart another round due to a lack of revision ideas and less guidance.

\subsubsection{Perception}

\textbf{Users with more AI-assistance are \textit{more positive agreement on the system capability to spark their creativity}, relative to those with less AI-assistance.} On a scale of 1 (very limiting to creativity) to 5 (very conductive to creativity), users with more AI-assistance (\varianttwo setting) reported a mean of 4.19 (SD=0.66) whereas users with less AI-assistance (\variantone) reported a mean of 3.58 (SD=0.79). The difference between these scores (0.61) is significant ($p=0.05$). This suggests that the inclusion of LLMs in the initial question forming or revision could also potentially boost the system's capacity to stimulate creativity. One user expressed similar thoughts: ``\textit{The ability to generate multiple iterations of questions using AI was helpful to my creativity, as well as the hints provided.}''

\textbf{\textit{Nine} users with both more and less AI-assistance report \textit{positive impressions} of the annotation system and task design.} 9 out of 27 users (5 from \variantone; 4 from \varianttwo) expressed their enthusiasm. One user from \variantone said ``\textit{I enjoyed learning how much the AI actually knew about culture,}''  Another user from \varianttwo liked the gamified elements: ``\textit{I liked how we could run the model again and again, ..., making this process a lot more fun and gamified.}''
Users from both settings positively referred to ``the confidence score'' helping them understand the effectiveness of their revisions.

\textbf{\textit{Some} users with both more and less AI-assistance \textit{enjoy} the LLM-powered modules.} 
Some users appreciate knowing the confidence score along with the chosen option by our Verifier (3 from \variantone, 1 from \varianttwo). This allows them to gauge how close they are to 'winning' (trick the AI successfully). Users with more AI-Assistance (6 from \varianttwo) enjoyed various LLM-powered assistance, emphasizing ``\textit{I liked how it generated a question based on your observation of a culture}''(4 from \varianttwo) and ``\textit{AI-generated hints} (2 from \varianttwo).''

One possible reason is that the LLM-powered modules reduce users' cognitive load when formulating questions and revisions. 
They provide grounded ideas based on users' inputs (e.g. their scenario, their posed question), rather than purely text guidance. Users engaged more by trying the LLM-powered modules, possibly increasing question revision time. 
When designing similar annotation systems, the trade-off between engagement and time spent should be considered.

\textbf{\textit{Six} users with both more and less AI-assistance report \textit{difficulty in deceiving the LLM}.} 
11 out of 27 users (6 from \variantone; 5 from \varianttwo) believed that nothing posed a challenge.  
However, 6 users (2 from \variantone; 4 from \varianttwo) expressed difficulty in deceiving the LLM: ``\textit{Refine the questions was a bit challenging. That might be due to the nature of the question rather than the platform itself.}''. Despite all our design modifications, such as incorporating a confidence score along with the response option, this remains difficult for some users, as evidenced by their feedback. This underscores the importance of designing novel annotation systems to assist human annotators in brainstorming scenarios rather than purely formulating questions.

\end{document}